\documentclass[twoside,11pt]{article}
\usepackage[preprint]{jmlr2e}
\usepackage[mode=buildnew]{standalone}
\usepackage[none]{hyphenat}
\usepackage{adjustbox}
\usepackage{amsmath}
\usepackage{array}
\usepackage{arydshln}
\usepackage{booktabs}
\usepackage{collcell}
\usepackage{datetime2}
\usepackage{etoolbox}
\usepackage{floatrow}
\usepackage{latexsym}
\usepackage{lipsum}
\usepackage{makecell}
\usepackage{multirow}
\usepackage{numprint}
\usepackage{pgfplots}
\usepackage{siunitx}
\usepackage{subcaption}
\usepackage{tcolorbox}
\usepackage{tikz}
\usepackage{times}
\usepackage{url}
\usepackage{verbatim}
\usepackage{xstring}
\usepackage{mathpazo}
\usepackage{xfp}

\usetikzlibrary{patterns, shapes.geometric, positioning, bayesnet}
\usepgfplotslibrary{fillbetween}

\newcommand{\slfrac}[2]{\left.#1\middle/#2\right.}
\newfloatcommand{capbtabbox}{table}[][\FBwidth]
\newcommand{\prob}[2]{
  \setlength{\thinmuskip}{2mu minus 1mu}
  \IfStrEq{#2}{}
    {p\left(#1\right)}
    {p\left(#1\middle|#2\right)}
  \setlength{\thinmuskip}{3mu}}

\newcommand*{\colorboxgap}{-1mm}    
\newcommand*{\colorboxwidth}{0.7cm}%
\newcommand*{\colorboxheight}{0.6cm}%

\newtoggle{inTableHeader}
\toggletrue{inTableHeader}
\newcommand*{\StartTableHeader}{\global\toggletrue{inTableHeader}}%
\newcommand*{\EndTableHeader}{\global\togglefalse{inTableHeader}}%
\newenvironment{colortabular}{
  \linespread{0.0}\selectfont
  \setlength{\tabcolsep}{0.7mm}
  \setlength{\fboxsep}{0mm}
  \StartTableHeader\tabular
}{
  \EndTableHeader\endtabular
}

\newcommand*{\MinNumber}{0.5}%
\newcommand*{\MidNumber}{0.75} %
\newcommand*{\MaxNumber}{1.0}%

\newcommand{\EmptyBox}[1]{%
  \vspace{\colorboxgap}
    \colorbox{white}{
      \parbox[c][\colorboxheight][c]{\colorboxwidth}{
      #1
    }}
}

\newcommand{\GradientBox}[2]{
  \ifdim #2 pt < 0pt
    \def\myminus{$\scalebox{0.5}[1.0]{\( - \)}$}
  \else
    \def\myminus{$\:\:$} 
  \fi
  \def\myabsval{\ifdim #2 pt < 0pt \fpeval{-1 * #2} \else #2 \fi}
  \vspace{\colorboxgap}
    \colorbox{#1}{
      \parbox[c][\colorboxheight][c]{\colorboxwidth}{
        \myminus\strut\num[round-mode=places,round-precision=2]{\myabsval}}
    }
}

\newcommand{\ApplyGradient}[1]{%
  \iftoggle{inTableHeader}{\EmptyBox{#1}}{
  \IfStrEq{#1}{ }{\EmptyBox{#1}}{
  \IfSubStr{#1}{---} {\EmptyBox{#1}} {
    \sisetup{add-integer-zero=false}
    \ifdim \MinNumber pt < \MaxNumber pt
        \ifdim #1 pt > \MidNumber pt
            \pgfmathsetmacro{\PercentColor}{max(min(100.0*(#1 - \MidNumber)/(\MaxNumber-\MidNumber),100.0),0.00)} %
            \GradientBox{green!\PercentColor!yellow}{#1}

        \else
            \pgfmathsetmacro{\PercentColor}{max(min(100.0*(\MidNumber - #1)/(\MidNumber-\MinNumber),100.0),0.00)} %
            \GradientBox{orange!\PercentColor!yellow}{#1}
        \fi
      \else
        \ifdim #1 pt > \MidNumber pt
            \pgfmathsetmacro{\PercentColor}{max(min(100.0*(#1 - \MidNumber)/(\MinNumber-\MidNumber),100.0),0.00)} %
            \GradientBox{orange!\PercentColor!yellow}{#1}
        \else
            \pgfmathsetmacro{\PercentColor}{max(min(100.0*(\MidNumber - #1)/(\MidNumber-\MaxNumber),100.0),0.00)} %
            \GradientBox{green!\PercentColor!yellow}{#1}
        \fi
      \fi
      \sisetup{add-integer-zero=true}
  }}}
}

\newcommand{\distcompare}[1]{
\begin{tikzpicture}
\begin{axis}[
    compat/BB=1.7,
    ymin=0, ymax=1, xmin=1, xmax=17,
    width=4cm, height=4cm,
    const plot, 
    axis lines=box, xtick=\empty, ytick=\empty,
    ]
\addplot+ [name path=one, 
            preaction={fill=blue},
            pattern=north east lines,
           thick, mark=none, opacity=0.2,
           ]
  table[x=indices,y=#1_0] {plots/trivial_dists_input.dat} \closedcycle;
\addplot+ [name path=two,
            preaction={fill=red},
            pattern=north west lines,
           thick, mark=none, opacity=0.2,
           ]
  table[x=indices,y=#1_1] {plots/trivial_dists_input.dat} \closedcycle;
\end{axis}
\end{tikzpicture}}

\newcolumntype{Y}{>{\collectcell\ApplyGradient}m{\colorboxwidth}<{\endcollectcell}}
\newcolumntype{W}[1]{>{\collectcell\EmptyBox}m{#1}<{\endcollectcell}}
\newcolumntype{R}[1]{>{\raggedleft\let\newline\\\arraybackslash\hspace{0pt}}m{#1}}
\newcolumntype{C}[1]{>{\centering\let\newline\\\arraybackslash\hspace{0pt}}m{#1}}
\newcolumntype{L}[1]{>{\raggedright\let\newline\\\arraybackslash\hspace{0pt}}m{#1}}
\newcolumntype{V}[1]{>{\adjustbox{angle=90,lap=\width-(#1)}\bgroup}l<{\egroup}}
\newcommand*\rot{\multicolumn{1}{V{1em}}}

\ShortHeadings{Generating Synthetic Text Data to Evaluate Causal Inference Methods}{Wood-Doughty, Shpitser and Dredze}
\firstpageno{1}

\begin{document}

\title{Generating Synthetic Text Data to Evaluate Causal Inference Methods}

\author{\name Zach Wood-Doughty \email zach@cs.jhu.edu \\
       \addr Johns Hopkins University\\
       Baltimore, MD 21218, USA
       \AND
       \name Ilya Shpitser \email ilyas@cs.jhu.edu \\
       \addr Johns Hopkins University\\
       Baltimore, MD 21218, USA
       \AND
       \name Mark Dredze \email mdredze@cs.jhu.edu \\
       \addr Johns Hopkins University\\
       Baltimore, MD 21218, USA
}

\editor{}

\maketitle

\begin{abstract}
Drawing causal conclusions from observational data requires making assumptions
about the true data-generating process. Causal inference research typically
considers low-dimensional data, such as categorical or numerical fields in
structured medical records. High-dimensional and unstructured data such as
natural language complicates the evaluation of causal inference methods; such
evaluations rely on synthetic datasets with known causal effects. Models for
natural language generation have been widely studied and perform well
empirically. However, existing methods not immediately applicable to
producing synthetic datasets for causal evaluations, as they do not allow for
quantifying a causal effect on the text itself. In this work, we develop a
framework for adapting existing generation models to produce synthetic text
datasets with known causal effects. We use this framework to perform an
empirical comparison of four recently-proposed methods for estimating causal
effects from text data. We release our code and synthetic
datasets.\footnote{\url{https://github.com/zachwooddoughty/causal_text_dgps}}
\end{abstract}

\section{Introduction}

Causal understanding is necessary for reasoning about hypothetical
interventions~\citep{pearl2018book}. As machine learning (ML) methods
demonstrate predictive success in complex domains, there is considerable
interest in relying on ML to make real-world decisions. However, predictive
models cannot be relied upon for decision-making without considering how
confounding or selection biases may affect the models'
predictions~\citep{char2018implementing,chen2017machine,liu2019number,subbaswamy2019preventing}.
Real-world interventions require causal reasoning, but causal reasoning requires
evaluations that go beyond traditional ML metrics such as test set accuracy. In
particular, causal methods rely on untestable assumptions about the
data-generating process (DGP) that produced the data. Violations of the
methods' assumptions may lead to biased predictions or estimates.

Researchers need complete knowledge of a DGP to test the assumptions of a
causal method, but such knowledge is often impossible for real-world datasets.
Thus while synthetic data has its
limitations~\citep{jensen2019comment,gentzel2019case}, it plays a crucial role
in understanding how a causal method performs when its assumptions are met or
violated. Recently, causal inference evaluations have tested proposed methods
against held-out synthetic
DGPs~\citep{hahn2019atlantic,dorie2019automated,shimoni2018benchmarking}. These
synthetic datasets are designed to test different empirical properties of the
methods, such as the coverage of confidence intervals or the finite-sample
behavior variance of an estimator.

Synthetic datasets are rarely used for predictive tasks when empirical
data is widely available. The enormous quantities of text and image data
have been curated to produce widely-used datasets for ML and natural language
processing (NLP) research~\citep{deng2009imagenet,brown2020language}.
However, synthetic datasets have been used in predictive tasks to explore how
models handle edge cases or low-resource
settings~\citep{elman1990finding,patki2016synthetic,khayrallah2018impact,
wang-eisner-2018-synthetic,kim-oneill-brown-2019-improving,winata-etal-2019-code}.
This is especially true in domains where data is not as widely available, such
as clinical settings~\citep{boag2016towards,belinkov2018synthetic,melamud2019towards}.

Causal methods have only recently been applied to natural language datasets.
\citet{keith2020text} provides a comprehensive overview of recent work,
focusing specifically on cases where text data can be used to adjust for
(otherwise unobserved) confounding. Text data provides a particularly difficult
domain for evaluating causal methods because it requires modeling causal
relationships between structured variables and text: ``what caused the author
to write the text this way?'' While there is plentiful text data for training
predictive models, we cannot directly measure the underlying processes that
humans use to produce or adapt their language in complex domains. Synthetic
DGPs need to balance `realism and control'~\citep{wendling2018comparing}:
the goal of producing realistic text data against the competing goal of
completely specifying the causal effects that produce the text. Past methods
evaluated on synthetic data have only satisfied one such goal, either by
producing particularly unrealistic text with known
effects~\citep{yao2019estimation,wood2018challenges,johansson2016learning} or
using real-world text without a fully-specified
DGP~\citep{veitch2020adapting,mozer2018matching,weld2020adjusting}.

We introduce a synthetic framework for evaluating causal methods that
incorporate text data, exploring desiderata of synthetic text DGPs and
tradeoffs between competing goals. We introduce two nontrivial synthetic DGPs,
one which samples a bag-of-words from an Latent Dirichlet Allocation (LDA) topic
model, and another which samples full sentences from
GPT-2~\citep{blei2003latent,radford2019language}. These two underlying
generative models allow us to test how causal methods perform when
their assumptions are violated (e.g. whether word order
matters)~\citep{wallach2006topic}. We use our framework to compare four causal
methods that rely on text, addressing a known gap in empirical evaluation of
such methods~\citep{keith2020text}.  We explore how existing methods' empirical
performance depends on their assumptions and show that when the causal
estimator depends on a text classifier model, better classification accuracy of
that classifier does not necessarily imply better causal estimates.  We release
our code and synthetic datasets to facilitate further development and
evaluation of causal methods for language data.

\section{Clinical Notes: A Motivating Example}
We begin by motivating causal inference for text data through an example.
Free text notes in medical records contain information about
patients' histories, possible diagnoses, or patient-doctor
relationships~\citep{rajkomar2018scalable,mcveigh2016can}.
Importantly, such information often does not appear anywhere else in a patient's
medical record, and thus is inaccessible to retrospective causal analyses that
do not use the free text data~\citep{wu2013evaluation,rosenbloom2011data,zheng2011handling}.

In this domain, assumptions about the DGP correspond to assumptions about how
clinical notes are written. Unless we have the requisite domain expertise to
precisely model the style, vocabulary, and semantics in the true DGP, we must be
particularly conservative about the assumptions we make. Synthetic DGPs allow us
to test how a method performs when its assumptions are violated, which is
essential to understanding whether to trust a real-world application.
While empirical success on synthetic data does not guarantee similar
performance on real data, any proposed method to draw causal inferences from
medical notes should first be validated on synthetic datasets that can capture
at least some of the complexity of human language. The goal of this work is
the development of synthetic DGPs for language data which make it possible
to evaluate causal inference methods.

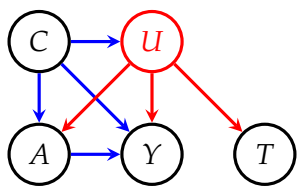
\begin{figure}[!t]
\centering
\begin{tikzpicture}[>=stealth, node distance=1.5cm]
    \tikzstyle{format} = [draw, very thick, circle, minimum size=0.8cm, inner sep=2pt]
    \tikzstyle{unobs} = [draw, very thick, red, circle, minimum size=0.8cm, inner sep=2pt]

    \begin{scope}
        \path[->, very thick]
            node[format] (A) {$A$}
            node[format, above of=A] (C) {$C$}
            node[unobs, right of=C] (U) {$U$}
            node[format, below of=U] (Y) {$Y$}
            node[format, right of=Y] (T) {$T$}

            (C) edge[blue] (A)
            (C) edge[blue] (Y)
            (C) edge[blue] (U)
            (U) edge[red] (A)
            (U) edge[red] (Y)
            (A) edge[blue] (Y)
            (U) edge[red] (T)
        ;

    \end{scope}
\end{tikzpicture}
\caption{The causal DAG we consider. $A$ is our treatment, $Y$ is our outcome,
$C$ and $U$ are confounders, and $T$ is the raw text which is influenced by
$U$. The counterfactual $p(Y(a))$ cannot be non-parametrically identified from
$\protect\prob{C, A, Y}{}$ alone due to unobserved confounding from $U$.
Methods may make parametric assumptions on the relationship between $T$ and $U$
in order to estimate the causal effect, or assume knowledge of
$\protect\prob{U}{T}$. We parameterize $\protect\prob{T}{U}$ with text
generation models in \S~\ref{sec:synthetic_dgps}.  We discuss the limitations
of this DAG model and extensions to other models in \S~\ref{subsec:other_dags}.}
\label{fig:dags}
\end{figure}

\section{Overview of Causal Assumptions} \label{sec:intro_causal}

While randomized control trials are the gold standard for determining causal
effects, they are often unethical, impossible, or prohibitively expensive.
Causal methods use non-randomized, observational data and assumptions about the
DGP to draw conclusions about hypothetical interventions. The ability to
make causal conclusions from observational data is transformative, but comes at
a cost. The methods require assumptions about the underlying DGP, and violation
of these assumptions can invalidate the model's conclusions. These assumptions
are often represented by a directed acyclic
graph~\citep[DAG;][]{pearl2009causality} like Figure~\ref{fig:dags}.

Imagine we want to study whether maternal vitamin D deficiency is a risk factor
for the pregnancy complication
preeclampsia~\citep{bodnar2014maternal,silva2008low}. In Figure~\ref{fig:dags},
the treatment $A$ is a binary measure of vitamin D deficiency and the outcome
$Y$ is the onset of preeclampsia. $C$ and $U$, age above 35 years and
socioeconomic status (SES), are confounders that influence both $A$ and $Y$.
Suppose SES is not directly recorded in structured (i.e. tabular) records, but
can be inferred from physician's text notes about the patient. While for
simplicity we will assume $A, C, U,$ and $Y$ are binary variables, we let $T$
denote the raw text of the clinical notes. The edge from $U$ to $T$ assumes
that the clinician's note-taking is influenced by the underlying $U$ value; the
lack of edges between $\{A, C, Y\}$ and $T$ reflects a simplifying assumption.
The relationship between $U$ and $T$ is complex and essential to the methods we
will consider.

In this setting, the target of interest is the average treatment effect;
how much more likely, on average, would patients suffer preeclampsia if they
were to have a vitamin D deficiency. We write this as $E[Y(1)] - E[Y(0)]$
where $Y(1)$ is a counterfactual random variable representing ``preeclampsia
status if a patient, possibly contrary to fact, had a vitamin D deficiency.''
This counterfactual variable's distribution can be identified as:
\begin{align}
p(Y(a)) = \sum_{C, U} \prob{Y}{A=a, C, U}\prob{C, U}{} \label{eq:g}
\end{align}
All confounders (common causes) must be included in Eq. (\ref{eq:g}) to draw
valid causal inferences \citep{pearl2009causality}. If we have no information
on $U$ and only observe $\prob{C, A, Y}{} = \sum_U \prob{Y, A, C, U}{}$, it is
generally impossible to write $p(Y(a))$ as a function of the observed
data~\citep{pearl2009causality}. In this case, we say $p(Y(a))$ is {\it not
identified}; it is impossible to derive a consistent estimator for the causal
effect. In real-world applications, an estimator for an unidentified effect may
return arbitrarily bad estimates. For a known DAG model, we can use the {\tt
ID} algorithm to determine whether a causal effect is identified given which
variables are observed~\citep{shpitser2006identification}.

In Figure~\ref{fig:dags}, we need nontrivial assumptions to identify
$p(Y(a))$ from $\prob{Y, A, C, T}{}$. The joint $\prob{U, T}{}$ determines
whether identification is possible. If $U \perp T$ ($T$ provides no information
on $U$), the causal effect is not identified and no method will succeed; if $T$
is an exact copy of $U$, then it should be trivial to recover the causal effect
by replacing $U$ with $T$ in Eq. (\ref{eq:g}). When $T$ is not an exact copy of
$U$, we may be able to treat it as a noisy, high-dimensional proxy for the
unobserved confounder $U$. Depending on the empirical relationship between the
text and the structured variables, methods that observe $T$ instead of $U$ may
be biased.

For real-world data, we cannot validate assumptions about the DGP. Therefore,
while applying a causal method to the data will produce conclusions given our
assumptions, it cannot validate the efficacy of the method itself. This is the
role of the synthetic DGP; we can compare the method's assumptions to a known
ground truth to explore how causal methods succeed or fail as the relationship
between text and structured data varies.

\begin{figure}[!t]
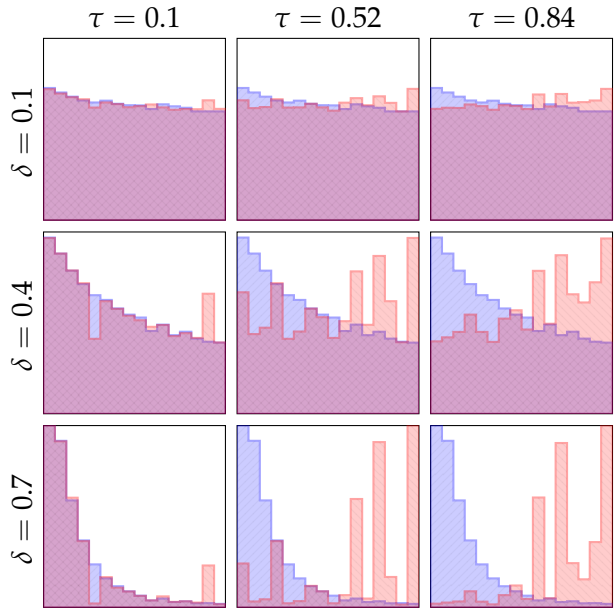

\centering
\setlength\tabcolsep{2pt}
\hspace{-1em}
\begin{tabular}{c c c c}
 & $\tau = 0.1$ & $\tau = 0.52$ & $\tau=0.84$ \\ 
\rot{\hspace{1.5em}$\delta=0.1$}
& \distcompare{0.1_0.1} & \distcompare{0.52_0.1} & \distcompare{0.84_0.1} \\
\rot{\hspace{1.5em}$\delta=0.4$}
& \distcompare{0.1_0.4} & \distcompare{0.52_0.4} & \distcompare{0.84_0.4} \\
\rot{\hspace{1.5em}$\delta=0.7$}
& \distcompare{0.1_0.7} & \distcompare{0.52_0.7} & \distcompare{0.84_0.7}
\end{tabular}
\caption{
Causal effect strengths and Trivial text generation. Blue and red bars
correspond to $U=0$ and $U=1$ respectively.  As $\tau$ increases, the ranked
preferences between $U=0$ and $U=1$ diverge.  As $\delta$ increases, the
distribution is puts more weight on the ranked preferences.  The x-axis indexes
the 16 words in the vocabulary, with each bar indicating the probability that a
word shows up at least once in a 16 word sequence.  When $\tau=0.1$ and
$\delta=0.1$, the distributions are close to uniform and almost entirely
overlap.  In all plots the $\tilde V_0$ order matches the x-axis order.  As
$\tau$ increases, the $\tilde V_1$ order diverges.  As $\delta$ increases, both
distributions become more concentrated on higher-ranked words.}
  \label{fig:trivial_plots}
\end{figure}

\section{Causal Effects in Text Generation} \label{sec:synthetic_dgps}

Recent work in natural language generation has introduced language models
with enormous empirical gains in perplexity and according to human
judgments~\citep{radford2019language,hashimoto2019unifying,brown2020language}.
Language models generate text sequences token by token, where token $i$
is sampled conditional on the previous $i-1$ tokens and the first token
is often sampled conditional on some initial context.
These existing methods, however, do not produce datasets with
known causal effects on text itself; we must first produce a formal
definition for the causal effect of a structured variable on the
text generation process. In our clinical
example, such an effect represents how a doctor's notes would
have changed had a patient, {\it counterfactually}, been of high SES. By
controlling the effect of $U$ on $T$ in Figure~\ref{fig:dags}, we can
evaluate how causal methods perform when their assumptions are met or
violated.

We want our marginal $p(T)$ to conform to a language model that generates text
according to a learned distribution, but want to parameterize $\prob{T}{U}$
such that we can force the generation to smoothly diverge from its learned
distribution to depend on $U$. We want a causal effect of $U$ on $T$ to make
some words or topics more likely and others less so.  That is, texts generated
when $U=1$ should be quantitatively and qualitatively different from texts when
$U=0$.  We will introduce $\tau$ and $\delta$ as hyperparameters that control
our causal effects. Intuitively, $\tau$ controls rankings over the vocabulary;
the larger $\tau$ is, the more the ranked preference for $U=0$ differs from
that of $U=1$.  We can conceptualize $\delta$ as controlling how much the model
indulges its preference; the larger $\delta$ is, the more likely
$\prob{T}{U=u}$ samples according to these ranked preferences rather than from the
pre-trained language model distribution.

\begin{table*}[!t]
  \centering
  \footnotesize
  \begin{colortabular}{W{1.0cm} *{3}{Y}}
  \toprule
  $\tau_\text{word}$ & 0.1 & 0.52 & 0.84   \\
  \cmidrule{1-1}
  $\delta_\text{word}$ & & & \EndTableHeader \\  
  0.1 & 0.544   & 0.596   & 0.623  \\
  0.4 & 0.752   & 0.965   & 0.979  \\
  0.7 & 0.780   & 1.000   & 1.000  \\
  \end{colortabular}
  \caption{$\protect\prob{U}{T}$ classifier accuracy for the nine examples of our trivial DGP.
  When both $\tau$ and $\delta$ are small, accuracy is near
  random chance. If one of the two parameters is large, accuracy improves;
  if both are large, accuracy nears 100\%}.
  \label{tbl:trivial_tc}
\end{table*}

To formalize this, let $V = \{x_1, \ldots, x_N\}$ be a vocabulary of $N$ words.
The learned language model provides an initial distribution $p(V)$ and uses it
to generate the sequences that comprise $p(T)$.
Let $\tilde V$ be an ordering over $V$ For a binary $U$, we choose two orderings,
$\tilde V_{u=1}$ and $\tilde V_{u=0}$. Our $\tau$ parameter controls the
correlation between those two orderings. When $\tau=0$, the orderings are
the same; when $\tau=1$, they are exact reversals of each other.
For a given $\tau$, we sample these orderings such that their Kendall Tau
correlation is approximately $1 - 2\tau$.

\begin{sloppy}
\binoppenalty=3000
\relpenalty=3000
For a given $\tilde V$ and our choice of $\delta$, we will construct a new
distribution over the vocabulary. Define $f_{\tilde V}(x_i)$ as a mapping from
a vocabulary item $x_i$ to the position of that item in the ordering. If
$x_{42}$ is the first item in the $\tilde V$ ordering, then
$f_{\tilde V}(x_{42}) = 1$. Now define a `modified Zipfian distribution'
as $p(x_i) \propto f_{\tilde V}(x_i)^{\slfrac{-\delta}{(1 - \delta)}}$,
When $\delta=0$, this is simply a uniform
distribution over the vocabulary; when $\delta=1$, it is a point mass on the
first item in its preference.\footnote{For any value of $\delta$, we will
normalize $p$ to be a distribution with probabilities in $[1e^{-10},
1 - 1e^{-10}]$.}
\end{sloppy}

Now, given our language model's learned $p(V)$, we construct a new
distribution:
\begin{align}
p'(x_i; \tilde V, \delta) &\propto p(x_i) \otimes f_{\tilde V}(x_i)^{-(\slfrac{\delta}{1 + \delta})}
\label{eq:effect}
\end{align}
where $\otimes$ indicates element-wise multiplication.
The distribution $p'$ represents an average between the initial
$p(V)$ and the modified Zipfian defined by $\tilde V$ and $\delta$.
We define a function $h$ which takes in an initial text generation distribution
$p(V)$, and values for $\tau$ and $\delta$ and returns new distributions
$p'_u$ following Eq. (\ref{eq:effect}).  We write this as:
\begin{align}
h: (p(V), \tau, \delta) \to \{p'_0(V; \tilde V_0, \delta), p'_1(V; \tilde V_1, \delta) \}
\label{eq:h_func}
\end{align}

\begin{table*}[!t]
\centering
\footnotesize
\begin{colortabular}{C{1.2cm} C{1.2cm} *{6}{Y}}
\toprule
      &  $\delta_\text{word}$ & 0.1   & 0.2   & 0.2   & 0.2   & 0.5   & 0.5   \\                  
      &  $\tau_\text{word}$     & 0.1   & .005 & 0.05  & 0.1   & .005 & 0.05 \\
\cmidrule{1-2}
$\delta_\text{topic}$     &  $\tau_\text{topic}$     &    &  &   &    &  & \EndTableHeader \\  
    0.1 &     0.1 & 0.561   & 0.546   & 0.584   & 0.643   & 0.628   & 0.944  \\
    0.2 &   0.005 & 0.558   & 0.542   & 0.586   & 0.643   & 0.629   & 0.944  \\
    0.2 &    0.05 & 0.626   & 0.619   & 0.650   & 0.695   & 0.688   & 0.947  \\
    0.2 &     0.1 & 0.654   & 0.647   & 0.672   & 0.716   & 0.727   & 0.953  \\
    0.5 &   0.005 & 0.653   & 0.638   & 0.676   & 0.730   & 0.713   & 0.963  \\
    0.5 &    0.05 & 0.873   & 0.861   & 0.877   & 0.887   & 0.910   & 0.989  \\
    0.5 &     0.1 & 0.966   & 0.966   & 0.969   & 0.973   & 0.980   & 0.993  \\
\end{colortabular}
\caption{$\protect\prob{U}{T}$ classification accuracy for LDA text.
Increasing $\tau$ and $\delta$ values lead to increased
classification accuracy, with exceptions when $\delta$ increases but $\tau$
decreases. If either the topic or word effects are particularly large,
classification accuracy exceeds $90\%$; when both are large, it quickly
approaches $100\%$.}
\label{tbl:lda_tc}
\end{table*}

Both $\tau$ and $\delta$ live in the $[0, 1]$ domain.
We can conceptualize $\tau$ as controlling the `preference' over words in the
vocabulary and $\delta$ as controlling the `strength' of that preference.
If either hyperparameter is $0$, the structured variable $U$ has no effect
on the text generation.
If $\tau=0$ then $\tilde V_1 = \tilde V_0$; while $\delta$ will change the word
probabilities, it will change them equally for either value of $U$.
Similarly, if $\delta$ is 0, then no matter how different $\tilde V_1$ is from
$\tilde V_0$, $h(p(V), \tau, 0)$ ignores those preferences and returns the
language model's learned $p(x_i)$. 

Figure~\ref{fig:trivial_plots} shows how $\delta$ and $\tau$ control a trivial
text generation model. We sample nine datasets of 10k sequences of 16 tokens.
Our initial $p(V)$ distribution is simply uniform over the vocabulary of 16 tokens.
Each cell in the figure shows how the Trivial $\prob{T}{U}$ distributions
change as we vary $\delta$ and $\tau$. When $\delta$ is large but $\tau$ is
small, some words are much more likely than others, but the two distributions
only differ on a single word. When $\tau$ is large but $\delta$ is small, the
distributions differ by a small amount on many words.

If we want to explore how causal methods perform in Figure~\ref{fig:dags},
we can control the $\prob{T}{U}$ distribution with $\delta$ and $\tau$.
As we turn to more complicated $p(V)$ distributions, we want a better way to
interpret the text generated with a given choice of these hyperparameters.

Our approach differs from past (semi-)synthetic text datasets for causal
evaluation. \citet{wood2018challenges} sampled synthetic `texts' in a 
bag-of-words manner similar to our Trivial distribution above, except without
the ability to control the strength of the $\prob{T}{U}$ relationship.
\citet{veitch2020adapting} used real text from Reddit or academic papers
and sampled synthetic outcomes conditional on metadata related to each text,
but without the ability to measure or specify the causal relationship
between the text and its metadata. \citet{weld2020adjusting} generate
semi-synthetic data by inserting template-based posts into the actual
post history of a social media user. These synthetic interventions are
discrete, however; there is no way to specify a real-valued causal effect and
manipulate it arbitrarily. The flexibility of our approach allows us to explore
how methods perform as we vary the causal effect on the text.

\subsection{Classification Accuracy and $\delta$, $\tau$}

The $\delta$ and $\tau$ hyperparameters completely control the effect of the
structured variables on the text, but are not particularly interpretable.
How do we know if particular $\delta$ or $\tau$ values are realistic? 
What values best mimic a real clinical notes DGP?

Rather than adapt our hyperparameters to a specific natural language domain,
we will use text classification accuracy as a lens that can be equally applied
to both synthetic and real-world text. Given a synthetic dataset,
we will train a classifier with $T$ as the features and $U$ as the labels.
Considering the accuracy of such a classifier will let us compare a synthetic
dataset to a real dataset; past work has extensively considered the task of
classifying clinical concepts from unstructured text~\citep{liu2018deep,%
meystre2008extracting,afzal2018natural,savova2010mayo}.
A synthetic dataset in which a text classifier achieves 99\% accuracy is
unrealistic, implying $\delta$ and $\tau$ are too large. Similarly,
if $\delta$ and $\tau$ are too small, a $\prob{U}{T}$ classifier will
be no better than chance.

\begin{table}[!t]
\centering
\footnotesize
\begin{colortabular}{C{3cm} *{11}Y}
\toprule
$\delta_\text{word}$ & 0.0 & 0.2 & 0.2 & 0.2 & 0.5 & 0.5 & 0.5 & 0.5 & 0.5 & 0.7 & 0.7 \\
$\tau_\text{word}$   & 0.00 & .025 & .025 & 0.15 & 0.05 & 0.05 & 0.05 & 0.15 & 0.15 & 0.15 & 0.15 \\
$\delta_\text{topic}$ & 0.7 & 0.7 & 0.7 & 0.7 & 0.5 & 0.7 & 0.7 & 0.7 & 0.9 & 0.7 & 0.9 \\
$\tau_\text{topic}$   & 0.45 & 0.15 & 0.45 & 0.15 & 0.05 & 0.05 & 0.45 & 0.05 & 0.15 & 0.15 & 0.15 \EndTableHeader \\
\midrule
Accuracy          & 0.619 & 0.544 & 0.639 & 0.622 & 0.647 & 0.642 & 0.698 & 0.854 & 0.854 & 0.776 & 0.793 \\
\bottomrule
\end{colortabular}
\caption{
$\protect\prob{U}{T}$ classification accuracy for GPT-2 text.
Accuracy is much lower than on trivial or LDA data.
Increasing $\tau$ and $\delta$ values generally leads to
increased classification accuracy, but this is not monotonic.
When increasing $(\delta_\text{word}, \tau_\text{word})$ from (0.5, 0.15) to
(0.7, 0.15) and reducing $(\delta_\text{template}, \tau_\text{template})$ from (0.9,
0.15) to (0.7, 0.15) we see a notable decrease in accuracy even when
$(\delta_\text{template}, \tau_\text{template})$ returns to (0.9, 0.15). This is
because GPT-2 word and template effects can conflict; because the language model
tries to maintain grammatical structure, certain templates make it unlikely
to sample certain words.
}
\label{tbl:gpt2_tc}
\end{table}

Table~\ref{tbl:trivial_tc} shows binary classification accuracy of a simple
bag-of-words model trained on the datasets from Figure~\ref{fig:trivial_plots}.
We use a train/dev/test split of 8k/1k/1k sequences for this and all
subsequent text classification experiments.
Accuracy improves above random chance as either $\delta$ or $\tau$ increase,
and quickly maxes out when both are large.
Classification accuracy on this task provides a useful way to abstract away the
underlying DGP as we introduce more complicated synthetic datasets.

\subsection{LDA with Causal Effects} \label{subsec:lda}

For a slightly more complicated synthetic DGP, we consider Latent Dirichlet
Analysis (LDA), one of the most widely-used models of
text~\citep{blei2003latent}. It provides a generative model of text that
clusters the distribution over the vocabulary into a distribution over topics.
While the LDA model ignores word order, so each sampled word drawn from the
trained model is independent. This results in generated texts that have no grammatical
structure.  We train an LDA model on a set of 250,000
documents which was released as part of the training data for
GPT-2~\citep{radford2019language}. 

To define a $\prob{T}{U}$ distribution that uses LDA, we will define causal
effects for both the words and the topics.
Let $V_\text{word}$ be the word vocabulary and $V_\text{topic}$ be the set of
learned topics. Then $p_\text{LDA}(V_\text{topic})$ is LDA's learned
baseline distribution over the topics,
and $p_\text{LDA}(V_\text{word} \mid t \in V_\text{topic})$ is the learned
distribution over words for topic $t$.

\begin{sloppy}
\binoppenalty=3000
\relpenalty=3000
We introduce causal effects with $h$ from (\ref{eq:h_func}). To sample a
word from our modified LDA model when $U=u$, we first sample a topic $t$ from
$h(p_\text{LDA}(V_\text{topic}), \tau_\text{topic}, \delta_\text{topic})$.
Then, instead of sampling from the original LDA distribution,
$p_\text{LDA}(V_\text{word} \mid t)$, we sample from
$h(p_\text{LDA}(V_\text{word} \mid t), \tau_\text{word}, \delta_\text{word})$.
\end{sloppy}

\begin{figure*}[!t]
\footnotesize
\begin{tabular}{cl}
\toprule
$\delta_\text{w}$ & The child was known for $\ldots$ \\
\midrule
0.0    & his role in the very real Peter Pan film that skyrocketed \\
0.1    & his role in the flamboyant sleuth Jackie Turner's hit \\
0.15   & his German business, and her books were sold in Bavaria \\ %
0.25   & her ability to play, run and shoot gags involving giant \\ 
0.4    & her ability to see. She began training one more spring \\ 
0.45   & her ability to see in one eye; her ability conquer magic \\ 
0.5    & her ability to disown her magic ability and her identify \\ 
0.6    & her ability one ability her magic ability her magic \\
\bottomrule
\end{tabular}
\caption{DistilGPT-2 generation when we fix the random seed, template, and
$\tilde V_\text{word}$ but vary $\delta_\text{word}$. We construct
$\tilde V_\text{word}$ so the most-preferred words are {\it her}, {\it magic},
and {\it ability}. The model switches from {\it his} to {\it her} pronouns as
$\delta$ increases. As $\delta$ further increases, sentence fluency decreases.}
\label{fig:gpt2_word_delta}
\end{figure*}

How do these $\tau$ and $\delta$ hyperparameters control the generated text?
Table~\ref{tbl:lda_tc} shows text classification results. We see that in
general, larger $\tau$ and $\delta$ lead to higher accuracy, yet there are
exceptions. Within a given row or column, when $\delta$ increases but $\tau$
decreases, we see a brief drop in accuracy. We can conceptualize this with the
plots in Figure~\ref{fig:trivial_plots}; as $\delta$ increases the effect of
$U$ on $T$ grows and the word distribution changes from its learned
distribution, but as $\tau$ decreases it decreases the difference between the
$U=0$ and $U=1$ `preference' distributions. If we plot $\tau_\text{word}$
against $\delta_\text{word}$ and hold topic effects constant, we would see that
accuracy monotonically increases as either word effect hyperparameter
increases.

\subsection{GPT-2 with Causal Effects} \label{subsec:gpt2}
One of the primary drawbacks of LDA is that it only models topic, and has no
sense of word order or syntax. Therefore, we consider a more complex DGP by
extending our synthetic data framework to more complicated neural models that
are widely used for text generation. 

GPT-2 is a large neural language model that has improved the state-of-the-art
on several benchmark evaluations~\citep{radford2019language}. It uses
1.5-billion parameters to encode a context sentence into an internal
representation and then uses that representation to predict a distribution over
the next word in the sentence. Once a word has been sampled from that
distribution, it is fed back into the model as additional context, and the
sampling process continues. Word-order is thus intrinsic to the sentences
generated by GPT-2. To save computation time, we use a smaller 82M parameter
DistilGPT-2 model~\citep{sanh2019distilbert}. We discuss extensions to more
recent neural language models in \S~\ref{subsec:newer_lms}.

While the model can take as input an arbitrary context sentence or phrase, we
follow \citet{sheng2019woman} and use a set of simple templates to seed
the generation of the GPT-2 model. The templates are a
combination of a subject (e.g. `the person') and the beginning of a verb phrase
(e.g. `was known for'). Our $V_\text{template}$ has 60 templates.
We treat GPT-2 as a black-box which inputs a distribution over
these 60 templates and outputs a distribution over the words in the
vocabulary. As with our LDA model, we will introduce causal effects which
influence these inputs and outputs, but otherwise leave the model untouched.

\begin{sloppy}
\binoppenalty=3000
\relpenalty=3000
We start with an initial uniform distribution over the 60 templates. From an
initially uniform $p_\text{GPT-2}(V_\text{template})$, we sample a template $t$
from $h(p_\text{GPT-2}(V_\text{template}), \tau_\text{template},
\delta_\text{template})$. Then, we feed that template into the GPT-2 model as
context, and it produces a distribution over words:
$p_\text{GPT-2}(V_\text{word} \mid t)$. We then sample the first word from
$h(p_\text{GPT-2}(V_\text{word} \mid t), \tau_\text{word}, \delta_\text{word})$.
We then feed that sampled word, $w_1$, back into the GPT-2 model and sample the
next word, conditioning on both the template and the first sampled word, from
$h(p_\text{GPT-2}(V_\text{word} \mid w_1, t), \tau_\text{word},
\delta_\text{word})$.
\end{sloppy}

Table~\ref{tbl:gpt2_tc} shows how text classification accuracy changes as our
$\tau$ and $\delta$ parameters change. As in Table~\ref{tbl:lda_tc}, larger
$\tau$ and $\delta$ values lead to better classification accuracy, with some
exceptions.
Every $\prob{U}{T}$ accuracy drop on LDA data in Table~\ref{tbl:lda_tc}
co-occurred with a drop in a $\tau$ or $\delta$ effect. With GPT-2,
we see one case where causal effects strictly increase but text classification
accuracy decreases. When $\tau_\text{word} = 0.15$ and $\delta_\text{template} = 0.7$
and $\delta_\text{word}$ increases from 0.5 to 0.7 and $\tau_\text{template}$ increases
from 0.05 to 0.15, text classification accuracy drops from 85\% to 78\%.
A likely explanation for this is that the GPT-2 templates do not affect
individual word probabilities, but provide context that affects the entire
sequence. The template fragment `worked as a' likely increases
occupation-related words, where the fragment `was known for' may not. These
non-monotonic effects may complicate the ability of our simple bag-of-words
model to differentiate the two distributions.

We also see that while the formal definitions of $\tau$ and $\delta$ are the same
between LDA and GPT-2, the values must be much larger for the classifier
to reach 90\% test set accuracy. This reflects the mismatch between the bag-of-words
assumption of our text classifier and the more complex text sequences of GPT-2.

As GPT-2 produces more fluent text than LDA, we can also visualize the effect of
$\delta_\text{word}$ by slightly varying its value while repeatedly sampling
from the model. Figure~\ref{fig:gpt2_word_delta} shows how the generation
changes when we fix the template and GPT-2's random seeds, and increase
$\delta_\text{word}$ for a given $\tilde V_\text{word}$ preference.

\section{Causal Methods with Text}

We have introduced a framework for producing datasets where we can provide
fine-grained control over how structured variables influence the text. We can
use this framework to evaluate existing methods for estimating causal effects
with text data. We will first provide an overview of four such
approaches, and then use our framework to conduct a range of simulation studies
that explore how well these methods perform as we vary the $\prob{T}{U}$
relationship.

Each method relies on sample-splitting for robust
inference~\citep{chernozhukov2016double,anderson2017split}. In particular,
we will split dataset in half, use one split to train and validate a simple
bag-of-words logistic regression model, and then use the other split to
estimate our causal effect. Then we will flip the splits to get a second
effect estimate on the first split, and then report the average of the two.
As we only use simple models for these evaluations, we leave full
implementation and training details to our released code.

\subsection{Matching with Text}

Matching is a popular causal method~\citep{stuart2010matching},
which has been recently applied to text
datasets \citep{roberts2018adjusting,mozer2018matching,yao2019estimation,wang2019words}.
Matching adjusts for confounding by estimating the causal effect among
patients who are similar, where similarity can be defined by confounders
or by their propensity to have received the treatment.
We consider two types of text matching: propensity score matching and representation matching.

If $U$ were observed in Figure~\ref{fig:dags}, valid propensity score matching
would proceed by learning a model for $\prob{A}{C,U}$ and matching patients
based on the estimated propensity. With $U$ unobserved, we will instead match
on a propensity score modeled as $\prob{A}{C,T}$. This method will be biased in
general because matching requires the true propensity score.  However, if there
exists a function that maps our estimated $\prob{A}{C, T}$ to the true
propensity $\prob{A}{C,U}$, this approach can be unbiased.  To implement this
method, we model the propensity $\prob{A}{C,T}$ with a bag-of-words classifier.
We then match on the estimated propensity using full matching as implemented in
the \texttt{R} package \texttt{optmatch}, following~\citet{mozer2018matching}.

Representation matching attempts to adjust for confounding by matching patients
on their covariates ($C$, $U$) and then taking $\prob{Y}{A}$ within each
matched group as an unbiased estimate of $\prob{Y(a)}{}$.  As $U$ is
unobserved, we can instead match on both $C$ and a learned representation of
$T$.  The intuition is that if two patients have similar $T$ representations,
they are likely to have the same value of $U$. However, this method will be
biased in general if two values $U$ can produce the same $T$ representation.
For our experiments, we use an LDA topic model representation of $T$ and
perform full matching using cosine similarity,
following~\citep{mozer2018matching}.

\subsection{Conditioning on Text}
\label{subsec:representations}

Rather than matching on the propensity score, we can directly use it in an
inverse propensity weighting (IPW) model~\citep{rosenbaum1983central}. This
approach reweighs the observed data by the inverse of the true propensity
model; if the true propensity $\prob{A}{C, U}$ is used, this is a consistent
estimator for Eq.~(\ref{eq:g}). When we replace $\prob{A}{C, U}$ with
$\prob{A}{C,T}$, our estimates are no longer guaranteed to converge to the
ground truth. Instead, we must assume that if the effect of $U$ on $T$ is
strong, then the learned propensity score will suffice to reweigh the examples.
This approach is similar to the bag-of-words method used
by~\citet{veitch2020adapting}. Initial experiments, we found that more powerful
neural models performed poorly on our datasets of only 10k examples. This
method follows other work in controlling for high-dimensional
confounders~\citep{hill2011challenges,mccaffrey2004propensity,low2016comparing}.

Our implementation again models $\prob{A}{C,T}$ as a
bag-of-words classifier. We truncate propensity weights and report
the mean of 100 bootstrap estimates~\citep{lee2011weight}. 

\subsection{Imputing with Text} \label{subsec:id_measurement_error}

Our fourth causal method assumes access to a text classifier model $\prob{U}{T}$
that can impute $U^*$, a noisy proxy for the true $U$. The method uses the classifier
and an estimate of the error rate of the classifier to correct for
the bias induced by the imperfect classifications~\citep{pearl2010measurement}.
Importantly, this approach requires more information than text matching or IPW,
as we must have access to either a pre-trained classifier with known error rate
or enough labeled data $p(U, T)$ to train a classifier. In many cases, such labeled data
may be difficult or impossible to collect.
We train a logistic regression classifier for $\prob{U}{T}$, using half the
training split to train the classifier, and the other half to estimate the
classifier's error rates. Our implementation uses code released
by~\citet{wood2018challenges}.

\begin{table*}[!t]
\footnotesize
\renewcommand{\MinNumber}{0.3}
\renewcommand{\MidNumber}{0.1}
\renewcommand{\MaxNumber}{0.0}
\newcommand*{\halfcolorboxheight}{0.3cm}%
\centering
\begin{colortabular}{C{0.8cm} *{3}{Y} @{\hskip \colorboxwidth} *{3}{Y} @{\hskip \colorboxwidth} *{3}{Y} @{\hskip \colorboxwidth} *{3}{Y}}
\toprule
& \multicolumn{3}{c}{\vrule width -\colorboxwidth height \halfcolorboxheight depth \halfcolorboxheight Representation}
& \multicolumn{3}{c}{\vrule width -\colorboxwidth height \halfcolorboxheight depth \halfcolorboxheight Propensity}
& \multicolumn{3}{c}{\vrule width -\colorboxwidth height \halfcolorboxheight depth \halfcolorboxheight IPW}
& \multicolumn{3}{c}{\vrule width 0pt height \halfcolorboxheight depth \halfcolorboxheight Measurement} \\
$\tau_\text{word}$ & 0.1 & 0.52 & 0.84  & 0.1 & 0.52 & 0.84 & 0.1 & 0.52 & 0.84  & 0.1 & 0.52 & 0.84 \\
\cmidrule(lr{\colorboxwidth}){2-4}  \cmidrule(lr{\colorboxwidth}){5-7}  \cmidrule(lr{\colorboxwidth}){8-10}  \cmidrule(lr){11-13}
$\delta_\text{word}$ & & & & & & & & & & & &\EndTableHeader \\  
    0.1 & 0.193   & 0.192   & 0.190  & 0.169   & 0.163   & 0.159  & 0.192   & 0.182   & 0.179 &  0.109   & 0.028   & 0.034  \\
    0.4 & 0.182   & 0.029   & 0.015  & 0.141   & 0.049   & 0.051  & 0.165   & 0.058   & 0.044 &  0.029   & 0.006   & 0.004  \\
    0.7 & 0.163   & 0.008   & 0.009  & 0.117   & 0.049   & 0.050  & 0.137   & 0.011   & 0.011 &  0.014   & 0.000   & 0.000  \\
\bottomrule
\end{colortabular}
\caption{Causal estimation error for the four estimation methods on our
trivial DGP. All methods approach zero error as $\delta$ and $\tau$ values increase.
}
\label{tbl:trivial_effects}
\end{table*}

\section{Evaluating Causal Methods with Text}

Our framework for producing synthetic text datasets
and discussed four past methods that have been proposed for estimating
causal effects from text datasets. We will now apply each of these four
methods -- text propensity score matching (Prop), text representation matching
(Rep.), IPW, and measurement error (ME) -- to the synthetic datasets we have
introduced. Our released code reproduces these experiments.

\subsection{Structured Variable Distribution} \label{sec:nontext-dgp}

In \S\ref{sec:synthetic_dgps}, we introduced hyperparameters that control
the causal effect of a structured variable on a text generation model.
To build our datasets, we first define $p(Y, A, C, U)$
and then define the text distribution $\prob{T}{U}$.
We limit ourselves to the DAG in Figure~\ref{fig:dags}
and only consider binary structured variables.

We choose the parameters of $p(Y, A, C, U)$ randomly, subject to three
constraints. First, we ensure that the true distribution-level causal effect
(\ref{eq:g}) is equal to $0.1$; given $C$ and $U$, the treatment increases the
likelihood of the outcome by $0.1$. Second, we ensure that our dataset
exhibits Simpson's paradox: if we estimate (\ref{eq:g}) {\bf without
conditioning} on $U$, the causal effect should appear to be $-0.1$. This setup
ensures that methods that completely ignore $U$ and $T$ will fail to estimate the
causal effect. Finally, we ensure that $p(U=1) = 0.5$, which makes a
majority-guess strategy for inferring $U$ maximally uninformative.
These constraints allow for consistency across experimental evaluations;
each structured distribution should be comparable.

\subsection{Reproducibility of Experiments}

Because we have a complex method for producing our text distribution
$\prob{T}{U}$ and we enforce non-trivial constraints on $p(Y, A, C, U)$, we
carefully seed the random number generation required to produce these synthetic
distributions. In particular, our sampling of text distributions and structured
distributions are orthogonal. We consider four separate structured distributions
that meet our above constraints, which we reuse in our evaluations across all
three text distribution settings: the trivial 16-word vocabulary, the LDA
model, and the GPT-2 model.

All results in Tables~\ref{tbl:trivial_effects}, \ref{tbl:lda_effects}, and
\ref{tbl:gpt2_effects} show the absolute-value divergence of the methods'
estimates from an oracle with access to the full structured distribution $p(Y,
A, C, U)$. The causal estimate errors for a given $(\tau, \delta)$ pair
are averaged over the 16 synthetic distributions that combine our four
structured distributions and four text distributions.

\begin{table*}[!t]
\renewcommand{\MinNumber}{0.3}
\renewcommand{\MidNumber}{0.1}
\renewcommand{\MaxNumber}{0.0}
\begin{tabular}{c c}
Representation & Propensity\\
\footnotesize
\begin{colortabular}{C{1.2cm} C{1.2cm} *{4}{Y}}
\toprule
      &  $\delta_\text{word}$ & 0.1   & 0.2   & 0.2   & 0.5   \\                  
      &  $\tau_\text{word}$   & 0.1   & 0.05  & 0.1   & 0.05 \\
\cmidrule{1-2}
$\delta_\text{topic}$     &  $\tau_\text{topic}$   & &    &  & \EndTableHeader \\  
    0.1 &     0.1 & 0.194   & 0.194   & 0.197   & 0.159  \\
    0.2 &    0.05 & 0.194   & 0.197   & 0.192   & 0.164  \\
    0.2 &     0.1 & 0.190   & 0.194   & 0.196   & 0.159  \\
    0.5 &    0.05 & 0.145   & 0.142   & 0.140   & 0.115  \\
\end{colortabular} &
\footnotesize
\begin{colortabular}{C{1.2cm} C{1.2cm} *{4}{Y}}
\toprule
      &  $\delta_\text{word}$ & 0.1   & 0.2   & 0.2   & 0.5   \\                  
      &  $\tau_\text{word}$   & 0.1   & 0.05  & 0.1   & 0.05 \\
\cmidrule{1-2}
$\delta_\text{topic}$     &  $\tau_\text{topic}$   & &    &  & \EndTableHeader \\  
    0.1 &     0.1 & 0.168   & 0.168   & 0.166   & 0.102  \\
    0.2 &    0.05 & 0.164   & 0.164   & 0.160   & 0.100  \\
    0.2 &     0.1 & 0.161   & 0.161   & 0.157   & 0.097  \\
    0.5 &    0.05 & 0.111   & 0.110   & 0.111   & 0.067  \\
\end{colortabular} \\
\\
IPW  & Measurement\\
\footnotesize
\begin{colortabular}{C{1.2cm} C{1.2cm} *{4}{Y}}
\toprule
      &  $\delta_\text{word}$ & 0.1   & 0.2   & 0.2   & 0.5   \\                  
      &  $\tau_\text{word}$   & 0.1   & 0.05  & 0.1   & 0.05 \\
\cmidrule{1-2}
$\delta_\text{topic}$     &  $\tau_\text{topic}$   & &    &  & \EndTableHeader \\  
    0.1 &     0.1 & 0.446   & 0.383   & 0.396   & 0.221  \\
    0.2 &    0.05 & 0.342   & 0.393   & 0.437   & 0.198  \\
    0.2 &     0.1 & 0.355   & 0.424   & 0.341   & 0.222  \\
    0.5 &    0.05 & 0.269   & 0.235   & 0.285   & 0.163  \\
\end{colortabular} &
\footnotesize
\begin{colortabular}{C{1.2cm} C{1.2cm} *{4}{Y}}
\toprule
      &  $\delta_\text{word}$ & 0.1   & 0.2   & 0.2   & 0.5   \\                  
      &  $\tau_\text{word}$   & 0.1   & 0.05  & 0.1   & 0.05 \\
\cmidrule{1-2}
$\delta_\text{topic}$     &  $\tau_\text{topic}$   & &    &  & \EndTableHeader \\  
    0.1 &     0.1 & 0.069   & 0.033   & 0.021   & 0.007  \\
    0.2 &    0.05 & 0.041   & 0.036   & 0.024   & 0.004  \\
    0.2 &     0.1 & 0.034   & 0.026   & 0.030   & 0.008  \\
    0.5 &    0.05 & 0.012   & 0.005   & 0.009   & 0.004  \\
\end{colortabular} \\
\end{tabular}
\caption{
Estimation error for each causal method on LDA
synthetic data, averaged over the combination of four structured
distributions and four text distributions for each cell.
All methods reduce estimation error as the $\delta$ and $\tau$ effects
increase in strength, but only measurement error achieves near-zero error
for any effect strength.
}
\label{tbl:lda_effects}
\end{table*}

\subsection{Evaluation with Trivial Text}

Table~\ref{tbl:trivial_effects} shows how the four causal methods perform on
the trivial 16-word vocabulary dataset we introduced in
\S~\ref{sec:synthetic_dgps}. We see that when the $\prob{T}{U}$ relationship is
very weak ($\delta_\text{w} = 0.1, \tau_\text{w} = 0.1)$, all four methods
perform about as poorly as they would if they had ignored the text entirely. As
the $\prob{T}{U}$ relationship becomes stronger, all four methods improve. The
text matching and measurement error methods are able to perfectly estimate the
true causal effect when the effect of $U$ on $T$ becomes overwhelmingly strong.
The IPW method does worse, but does correct for the $U$ confounding as the
$\prob{T}{U}$ relationship strengthens. It is not surprising that the
measurement error approach works here, as Table \ref{tbl:trivial_tc} 
and Figure~\ref{fig:trivial_plots}
showed us that $\prob{U}{T}$ classification can achieve
perfect accuracy on this trivial dataset. The success of the text matching
approach highlights that even though $\prob{A}{C,T}$ is not the true propensity
score, the relationship between $U$ and $T$ is strong enough to
allow for the method to correct for the confounding.

\subsection{Evaluation with LDA Text}
Table~\ref{tbl:lda_effects} shows how the four causal methods perform on
synthetic datasets using the LDA text generation we introduce in
\S~\ref{subsec:lda}. These results are less encouraging. Our text
generated from LDA is word-order independent, so simple
bag-of-words models $\prob{A}{C,T}$ should be powerful enough to
capture the text's complexity. Even so, the matching methods struggle
to correct for $U$'s confounding, though they slightly improve as $\tau$ and
$\delta$ increase. Compared to the trivial setting, in LDA there is much
less direct relationship between $U$ and the sampled text. Thus Representation
matching is more likely to match two texts with different $U$ values, and
in Propensity the estimated $\prob{A}{C,T}$ diverges from the true propensity.
That Propensity outperforms Representation when it did not for Trivial text
suggests that the propensity matching may be more effective as it in a single
dimension~\citep{roberts2018adjusting}.
The IPW method, on the other hand, does extremely poorly when the effects of
$U$ on $T$ are small. Because a na\"ive estimator that ignores the text can
achieve a causal error of $0.20$, the IPW estimator actually worsens the
confounding bias. The measurement error approach is effective when $\tau$ and
$\delta$ are large enough.

\begin{table}[!t]
\renewcommand{\MinNumber}{0.3}
\renewcommand{\MidNumber}{0.1}
\renewcommand{\MaxNumber}{0.0}
\footnotesize
\centering
\begin{colortabular}{C{3cm} *{11}Y}
\toprule
$\delta_\text{word}$ & 0.0 & 0.2 & 0.2 & 0.2 & 0.5 & 0.5 & 0.5 & 0.5 & 0.5 & 0.7 & 0.7 \\
$\tau_\text{word}$   & 0.00 & .025 & .025 & 0.15 & 0.05 & 0.05 & 0.05 & 0.15 & 0.15 & 0.15 & 0.15 \\
$\delta_\text{template}$ & 0.7 & 0.7 & 0.7 & 0.7 & 0.5 & 0.7 & 0.7 & 0.7 & 0.9 & 0.7 & 0.9 \\
$\tau_\text{template}$   & 0.45 & 0.15 & 0.45 & 0.15 & 0.05 & 0.05 & 0.45 & 0.05 & 0.15 & 0.15 & 0.15 \EndTableHeader \\
\midrule
Representation & 0.193 & 0.197 & 0.193 & 0.194 & 0.192 & 0.193 & 0.190 & 0.181 & 0.180 & 0.172 & 0.163 \\
Propensity & 0.158 & 0.168 & 0.155 & 0.165 & 0.166 & 0.165 & 0.156 & 0.130 & 0.131 & 0.101 & 0.098 \\
IPW & 0.177 & 0.197 & 0.171 & 0.187 & 0.173 & 0.189 & 0.158 & 0.102 & 0.092 & 0.123 & 0.112 \\
Measurement & 0.101 & 0.104 & 0.042 & 0.026 & 0.026 & 0.035 & 0.031 & 0.012 & 0.015 & 0.013 & 0.014 \\
\bottomrule
\end{colortabular}
\caption{
Causal estimation error for each method on the GPT-2 synthetic data. The
measurement error method estimates approach zero only for the largest values
of $\delta$ and $\tau$. Neither Propensity nor IPW correct more than half the confounding
of a naive estimator, and Representation barely reduces the confounding bias at
all.}
\label{tbl:gpt2_effects}
\end{table}

\subsection{Evaluation with GPT-2 Text}

Table~\ref{tbl:lda_effects} shows how the four causal methods perform on
synthetic datasets using the GPT-2 text generation we introduce in
\S~\ref{subsec:gpt2}. Here we see that neither the matching nor IPW methods
ever noticeably improve. The measurement error method is still effective,
but only when the effect of $U$ on $T$ is strongest.

While GPT-2 clearly does not produce language at the complexity of real-world
datasets, we can better understand the assumptions made by these causal models
by exploring how they perform as the underlying text generation become more
complex. On this data, simple bag-of-words models we consider are not
flexible enough to fully capture the complexity of the text. Even though
Table~\ref{tbl:gpt2_tc} shows us that a bag-of-words classifier can effectively
learn this more complicated $\prob{U}{T}$ when the word and template effects
are large enough, the $\prob{A}{C,T}$ model learned for the IPW and matching
methods does not capture information on the true propensity. The measurement
error method and its $\prob{U}{T}$ classifier can provide unbiased estimates,
but only when $\delta$ and $\tau$ effects are strongest.

\begin{figure}[!t]
\centering
\includegraphics[width=\linewidth]{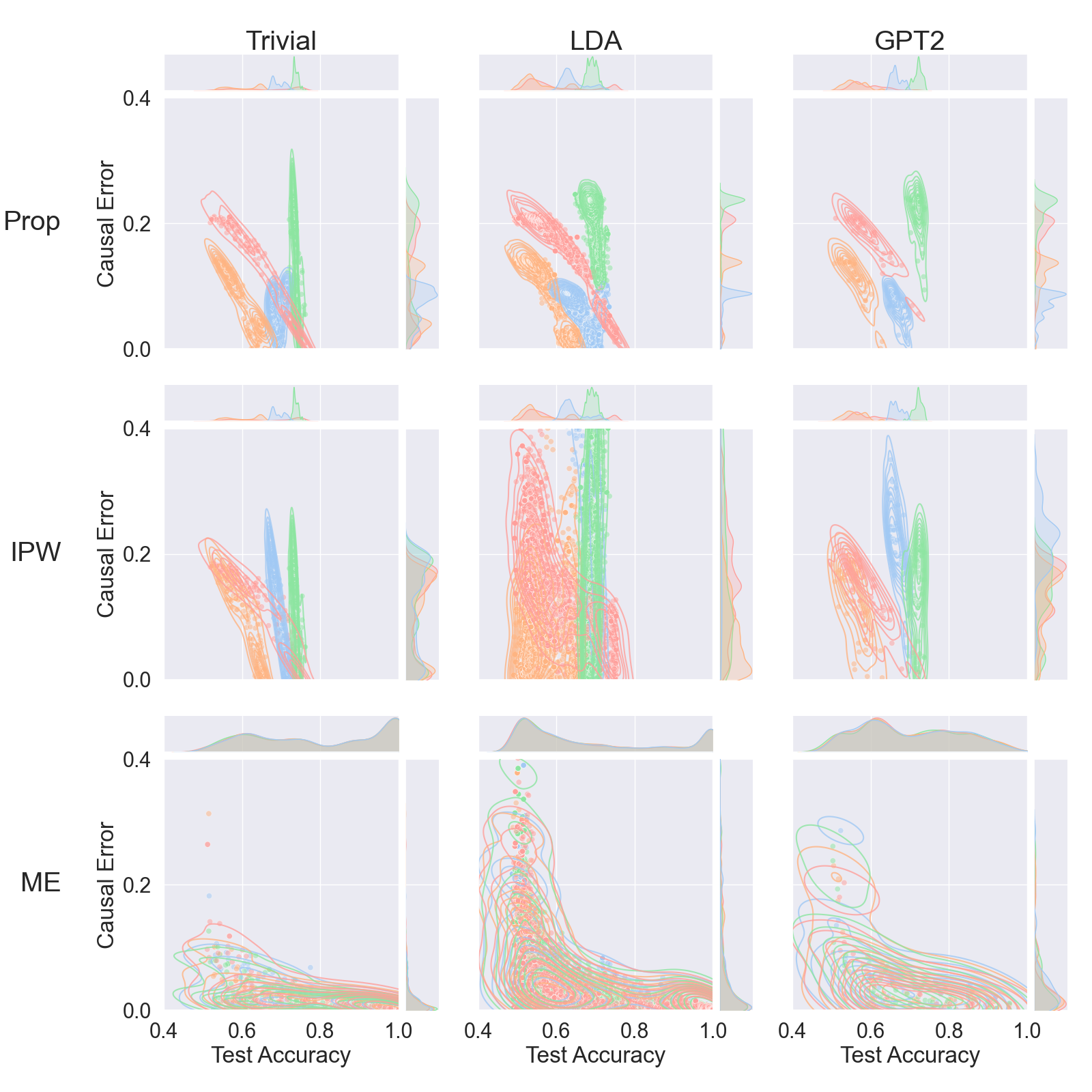}
\caption{
Joint and marginal density plots of text classifier accuracy and mean absolute causal estimation error
for each DGP and each estimation method that relies on a text classifier.
Each dot represents one experiment. Figure~\ref{fig:acc_vs_err_ipw_lda} shows
a zoomed-out plot for LDA+IPW; all other plots contain all data.
Colors indicate the four structured variable random seeds used to create the true
data-generating distributions.
For the IPW and Prop methods, the visible clusters show that
the relationship between classifier accuracy and causal error is highly
dependent on the random seed for structured variables. Thus, for a real-world analysis
with an unknown DGP, better classifier accuracy does not imply lower causal error.
For the ME method, classifier accuracy and causal error are not clustered
by the underlying DGP.
}
\label{fig:acc_vs_err}
\end{figure}

\begin{figure}[!t]
\centering
\includegraphics[width=0.5\linewidth]{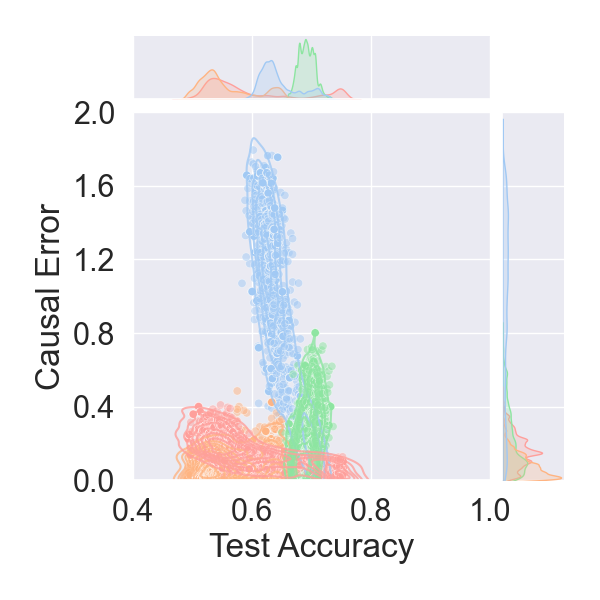}
\caption{
Zoomed-out version of Figure~\ref{fig:acc_vs_err} for
for IPW estimator on LDA data. For one random seed for
structured variables (the blue cluster), causal error
is quite large.
}
\label{fig:acc_vs_err_ipw_lda}
\end{figure}

\section{Text Classification Accuracy and Estimation Error}

\begin{table*}
\renewcommand{\MinNumber}{0.1}
\renewcommand{\MidNumber}{-0.1}
\renewcommand{\MaxNumber}{-0.6}
\footnotesize
\centering
\begin{colortabular}{C{3cm} *{3}Y}
\toprule
DGP & Prop & IPW & M.E. \EndTableHeader \\
\midrule
Trivial & -.234 & -.433 & -.567 \\
LDA     & -.139 &  0.058 & -.584 \\
GPT-2   & 0.107 &  0.011 & -.571 \\
\bottomrule
\end{colortabular}
\caption{Pearson correlation between absolute causal estimation error
and the test accuracy of the text classifier that the
estimation method relies on. On the Trivial text data,
all methods have a negative correlation: increased test accuracy
implies lower estimation error. As the text DGP increases in complexity to
LDA and GPT-2, this correlation dwindles and then reverses for the Propensity
and IPW methods, but remains stable for the measurement method.
}
\label{tbl:text_acc_error_correlations}
\end{table*}

Our propensity score matching, IPW, and measurement error methods all rely in
part upon a text classifier to estimate the causal effect. However, better
performance (as measured by classification accuracy) of this classifier does
not necessarily translate into lower causal estimation error. For both
propensity score matching and IPW, the text classifier models $\prob{A}{C,T}$.
For the measurement error estimator, the text classifier models $\prob{U}{T}$.
For the binary $A$ and $U$ we consider, we can easily characterize these models
in terms of their classification accuracy. The density plots in
Figure~\ref{fig:acc_vs_err} shows the relationship between text classifier
accuracy and the causal estimation error.

Across all three DGPs, we see that when the $\prob{U}{T}$ classifier has
accuracy greater than 80\%, our estimate of the causal effect is within
0.05 of the truth. If we could achieve 100\% classifier accuracy for the measurement
method, it would imply that we had access to the true $\prob{A, Y, C, U}{}$,
and can trivially estimate the causal effect. 

However, for propensity and IPW methods, better classification accuracy does
not imply lower estimation error. In fact, better classification accuracy of
$\prob{A}{C,T}$ is orthogonal to our goals of low causal estimation error.
Instead, we need $\prob{A}{C, T}$ to converge to the true $\prob{A}{C,U}$,
which is untestable without observing $U$. 

Table~\ref{tbl:text_acc_error_correlations} shows that as we increase the
complexity of our DGP from the Trivial text to LDA and then to GPT-2,
we can also empirically see that the correlation between classifier accuracy
and estimation error degrades for the Propensity and IPW methods.
For the Propensity and IPW methods on GPT-2 data, classifier accuracy
is positively correlated with estimation error, suggesting that the
$\prob{A}{C, T}$ classifier has overfit and diverged from the true
$\prob{A}{C, U}$ propensity.

\begin{table*}[!t]
\renewcommand{\MinNumber}{0.3}
\renewcommand{\MidNumber}{0.1}
\renewcommand{\MaxNumber}{0.0}
\footnotesize
\centering
\begin{colortabular}{C{1.4cm} C{1.4cm} C{1.4cm} C{1.4cm} *{11}Y }
\toprule
 & & & & \multicolumn{11}{c}{Labeled $\prob{U,T}{}$ examples} \\
\cmidrule{5-15}
$\delta_\text{w}$ & $\tau_\text{w}$ & $\delta_\text{t}$ & $\tau_\text{t}$ &
50 & 100 & 200 & 300 & 400 & 500 & 1000 & 1500 & 2000 & 2500 & 5000 \EndTableHeader   \\
\midrule
0.0 & 0.00 & 0.7 & 0.45 & 0.186 & 0.159 & 0.135 & 0.111 & 0.113 & 0.100 & 0.083 & 0.082 & 0.074 & 0.109 & 0.101 \\
0.2 & 0.03 & 0.7 & 0.15 & 0.185 & 0.184 & 0.157 & 0.173 & 0.162 & 0.160 & 0.157 & 0.137 & 0.103 & 0.103 & 0.104 \\
0.2 & 0.03 & 0.7 & 0.45 & 0.181 & 0.168 & 0.127 & 0.090 & 0.109 & 0.098 & 0.100 & 0.075 & 0.073 & 0.063 & 0.041 \\
0.2 & 0.15 & 0.7 & 0.15 & 0.186 & 0.171 & 0.131 & 0.148 & 0.128 & 0.090 & 0.055 & 0.041 & 0.034 & 0.030 & 0.026 \\
0.5 & 0.05 & 0.5 & 0.05 & 0.174 & 0.157 & 0.131 & 0.114 & 0.122 & 0.094 & 0.042 & 0.033 & 0.041 & 0.034 & 0.026 \\
0.5 & 0.05 & 0.7 & 0.05 & 0.183 & 0.161 & 0.140 & 0.132 & 0.131 & 0.122 & 0.057 & 0.030 & 0.040 & 0.040 & 0.035 \\
0.5 & 0.05 & 0.7 & 0.45 & 0.180 & 0.141 & 0.117 & 0.103 & 0.088 & 0.101 & 0.059 & 0.031 & 0.026 & 0.027 & 0.031 \\
0.5 & 0.15 & 0.7 & 0.05 & 0.099 & 0.053 & 0.062 & 0.041 & 0.037 & 0.023 & 0.021 & 0.014 & 0.013 & 0.013 & 0.011 \\
0.5 & 0.15 & 0.9 & 0.15 & 0.113 & 0.071 & 0.061 & 0.045 & 0.041 & 0.031 & 0.029 & 0.016 & 0.017 & 0.019 & 0.015 \\
0.7 & 0.15 & 0.7 & 0.15 & 0.099 & 0.075 & 0.051 & 0.035 & 0.035 & 0.029 & 0.018 & 0.021 & 0.019 & 0.018 & 0.013 \\
0.7 & 0.15 & 0.9 & 0.15 & 0.095 & 0.067 & 0.037 & 0.041 & 0.035 & 0.031 & 0.020 & 0.021 & 0.018 & 0.016 & 0.014 \\
\midrule
\multicolumn{4}{c}{$p(U, C, A, Y)$ Baseline}
& 0.282   & 0.136   & 0.048   & 0.025   & 0.040   & 0.030   & 0.028   & 0.015   & 0.009   & 0.008   & 0.007 \\
\bottomrule
\end{colortabular}
\caption{
Measurement error method's mean absolute estimation error on GPT-2 data as we
vary the amount of labeled data used. Train and validation data is split
evenly; we train the $\protect\prob{U}{T}$ classifier with half and estimate
its error rate on the other half. The last column is equivalent to the last row
of Table~\ref{tbl:gpt2_effects}. The $\protect\prob{U, C, A, Y}{}$ baseline
ignores the text and simply computes the causal effect using
Equation~\ref{eq:g}.
}
\label{tbl:measurement_n_labeled}
\end{table*}

\begin{figure}[!t]
\centering
\includestandalone[width=\linewidth,mode=image]{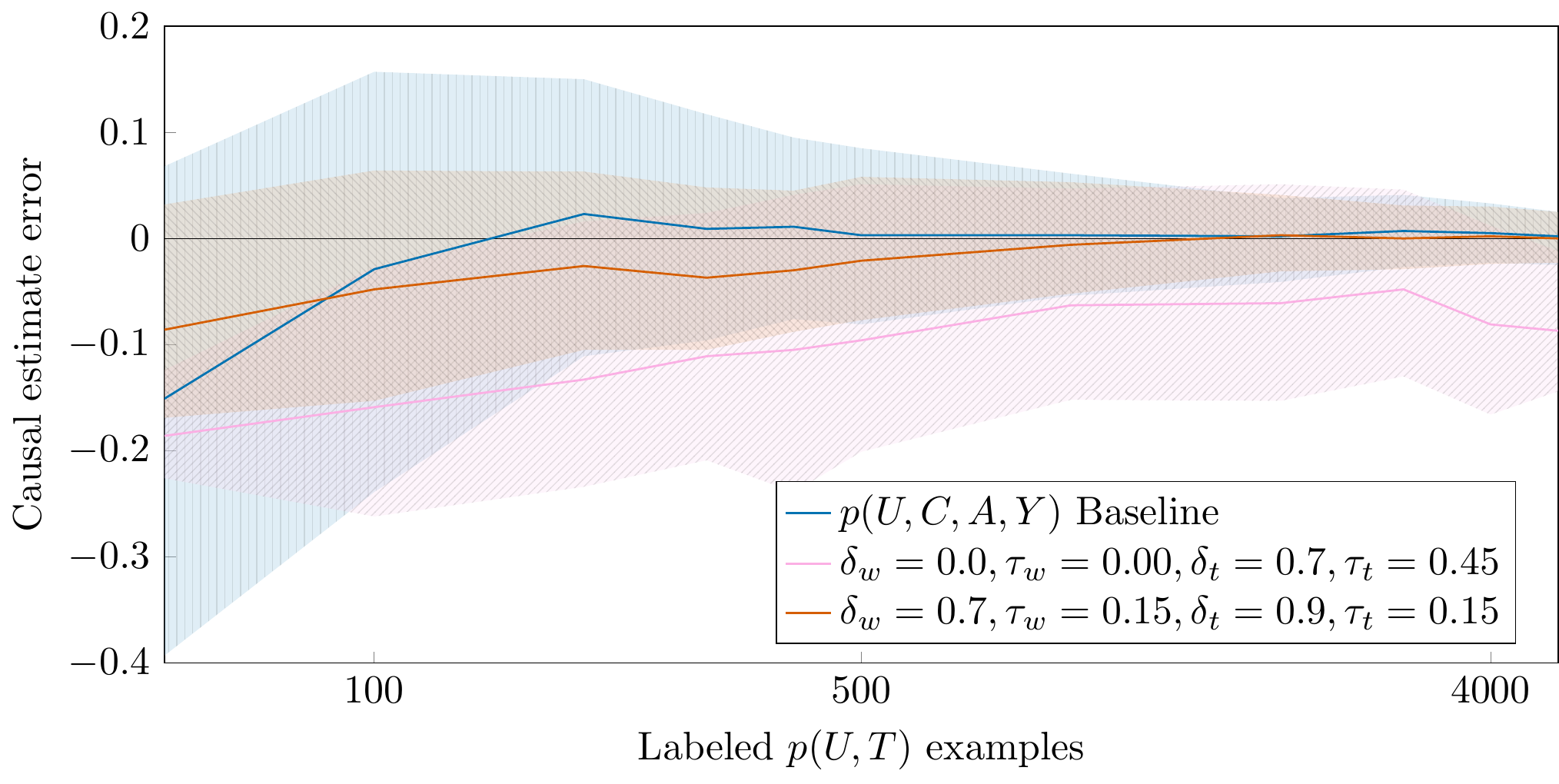}
\caption{A closer look at three rows from Table~\ref{tbl:measurement_n_labeled}.
Solid line plots mean (not mean absolute) causal error;
shaded regions show 95\% confidence interval from 100 bootstrap samples.
Measurement error results are averaged over four structured variables distributions and four
text distributions. The baseline ignores the text and is averaged over four structured distributions.
Even for text data with the strongest causal effects
we consider, the measurement error approach is not noticeably better than
the $\protect\prob{U, C, A, Y}{}$ baseline once we have at least 200 labeled examples.}
\label{fig:n_labeled_bounds}
\end{figure}

\section{Availability and Use of Labeled $U$ Data} \label{sec:n_labeled}

Our empirical results have demonstrated that the measurement error estimator
performs the best on our synthetic datasets. However, this method relies upon
access to labeled $\prob{U}{T}$ data. This finding raises two questions: how
much labeled data does the measurement error method require, and could other
methods perform as well or better if given access to such labeled
$\prob{U,T}{}$ data?

We run additional experiments where we limit the amount of labeled data that
our estimator has access to. Of a dataset of 10,000 total examples, we use $n$
of them to train and validate a classifier $\prob{U}{T}$ and use $(10,000 - n)$
to compute our estimate of the causal effect. Our previous experiments have
considered $n = 5,000$; in Table~\ref{tbl:measurement_n_labeled} we plot
estimation error as we vary $n$ from $50$ to $5,000$.  For the DGPs with the
strongest causal effects, the mean absolute error remains small even as we
substantially reduce the number of examples. Estimation error on DGPs with
weaker causal effects are more sensitive to the number of examples.

We then compare these evaluations on limited labeled data against a baseline
that assumes access to an equal amount of data on the full $\prob{U, C, A,
Y}{}$ distribution. Suppose we can pay clinicians to annotate $n$ patient
records for the unobserved confounder $U$; should we use those examples to use
the measurement error method, or should we just directly compute the causal
effect using Equation~(\ref{eq:g}), ignoring the text entirely? The
$\prob{U, C, A, Y}{}$ baseline in Table~\ref{tbl:measurement_n_labeled}
suggests that as soon as we have at least 200 examples, this baseline is as
good on average as the measurement error method, even on DGPs with the
strongest $U\to T$ causal effects. Figure~\ref{fig:n_labeled_bounds}
shows in more detail this baseline compared against two of the DGPs in
Table~\ref{tbl:measurement_n_labeled}. In particular, this figure shows
the 95\% confidence interval for the three methods. For the DGP with
large causal effects, the measurement error method is quite comparable
to the baseline as $n \geq 500$, but has somewhat smaller confidence
intervals at lower-data settings. On the DGP with small $U \to T$ causal effects,
the measurement error method is strictly worse than the baseline.

The measurement error method is the only approach that achieves success on our
GPT-2 DGPs, but requires access to $\prob{U}{T}$ labels. If this method can be
matched by a baseline that ignores the text entirely, it may seem that
incorporating NLP methods into causal inference is not worth the effort.  But
our results are not entirely pessimistic and the flaws they do reveal point to
many opportunities for future work. The $\prob{U, C, A, Y}{}$ baseline
importantly requires access to the full joint, whereas the measurement error
method only requires data on the $\prob{U}{T}$ conditional. This has many
practical implications. For example, if researchers at a hospital cannot
collect $U$ annotations for their data due to patient privacy restrictions,
they still may be able to apply a $\prob{U}{T}$ classifier to that data.  Thus
if we can leverage existing anonymized clinical datasets as the $\prob{U, T}{}$
data, we can produce analysis that would otherwise be impossible.

There are also many opportunities to develop new approaches that outperform the
four methods we evaluated.  We should expect that some access to labeled data
should make it possible to learn a propensity score or text representation that
provides for lower estimation error when primarily using data without labeled
$U$.  An unsupervised text representation such as LDA could be augmented with
labeled $\prob{U,T}{}$ so that learned topics are more discriminative of the
underlying $U$~\citep{blei2007supervised}. Similarly, if we were given access to
some labeled $\prob{U,T,C}{}$ data, we could train a propensity score model
such that predicted propensities must be roughly equal for examples with the
same $U$.
We can also explore approaches that combine these four methods to produce
new \emph{multiply-robust} methods. Many causal estimators use multiple models
and are provably unbiased if at least one or more of those models are
correctly-specified~\citep{bang2005doubly,vansteelandt2008multiply}. Can we
develop a new matching method that are unbiased if either the propensity model
\emph{or} representation model are unbiased?  Can we effectively combine all
four methods we considered into a single multiply-robust estimator?

\section{Limitations and Extensions}

Our evaluation framework and experimental results provide new insights
into how existing estimators perform on synthetic text datasets.
In generating our synthetic datasets and evaluating these methods,
we have made simplifying assumptions. Many of these assumptions may limit
the efficacy of our work to certain applications, yet most such assumptions
can be relaxed by extending our work.

\begin{figure}[!t]
\begin{subfigure}[t]{0.48\textwidth}
\centering
\captionsetup{width=0.8\textwidth}
\begin{tikzpicture}[>=stealth, node distance=1.5cm]
    \tikzstyle{format} = [draw, very thick, circle, minimum size=0.8cm, inner sep=2pt]
    \tikzstyle{unobs} = [draw, very thick, red, circle, minimum size=0.8cm, inner sep=2pt]

    \begin{scope}
        \path[->, very thick]
            node[format] (A) {$A$}
            node[format, above of=A] (C) {$C$}
            node[unobs, right of=C] (U) {$U$}
            node[format, below of=U] (Y) {$Y$}
            node[format, right of=Y] (T) {$T$}

            (C) edge[blue] (A)
            (C) edge[blue] (Y)
            (C) edge[blue] (U)
            (U) edge[red] (A)
            (U) edge[red] (Y)
            (A) edge[blue] (Y)
            (A) edge[blue, bend right=30] (T)
            (C) edge[blue, bend left=5] (T)
            (Y) edge[blue] (T)
            (U) edge[red] (T)
        ;

    \end{scope}
\end{tikzpicture}
\caption{A DAG in which all structured variables influence text generation.}
\label{fig:dag_differential}
\end{subfigure}
\begin{subfigure}[t]{0.48\textwidth}
\centering
\captionsetup{width=0.8\textwidth}
\begin{tikzpicture}[>=stealth, node distance=1.5cm]
    \tikzstyle{format} = [draw, very thick, circle, minimum size=0.8cm, inner sep=2pt]
    \tikzstyle{unobs} = [draw, very thick, red, circle, minimum size=0.8cm, inner sep=2pt]

    \begin{scope}
        \path[->, very thick]
            node[format] (A) {$A$}
            node[format, above of=A] (C) {$C$}
            node[unobs, right of=C] (U) {$U$}
            node[format, below of=U] (T) {$T$}
            node[format, right of=T] (Y) {$Y$}

            (C) edge[blue] (A)
            (C) edge[blue] (Y)
            (C) edge[blue] (U)
            (U) edge[red] (A)
            (U) edge[red] (Y)
            (A) edge[blue] (T)
            (C) edge[blue, bend left=5] (T)
            (T) edge[blue] (Y)
            (U) edge[red] (T)
        ;

    \end{scope}
\end{tikzpicture}
\caption{A DAG in which the text acts as either a treatment or outcome.}
\label{fig:text_treatment_outcome}
\end{subfigure}
\caption{Causal DAG models to which our evaluation framework could be extended.}
\label{fig:harder_dags}
\end{figure}
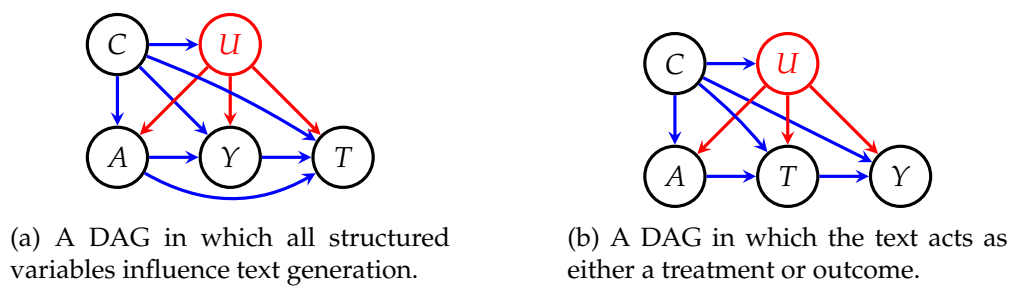

\subsection{Other DAG Models} \label{subsec:other_dags}

We only sample datasets from synthetic DGPs corresponding to the
DAG model in Figure~\ref{fig:dags}. There are of course infinitely
many DAG models that could be considered, but we point out a few
important generalizations that would complicate our methods for
sampling data or evaluating methods.

Figure~\ref{fig:dag_differential} extends Figure~\ref{fig:dags} by
adding causal effects from \emph{all} structured variables to the
text data. Such a DAG complicates our approach for sampling text
from a language model conditional on the structured variables.
In \S~\ref{sec:synthetic_dgps} we parameterized $\prob{T}{U}$ with our two
types of hyperparameters: $\delta$ and $\tau$.
Figure~\ref{fig:dag_differential} requires sampling from $\prob{T}{U, C, A,
Y}$, which may require a different hyperparameter formulation.
Our implementation assumes $U$ is binary, the immediate extension to a
continuous-valued $U$ simply requires replacing the two orderings ($\tilde
V_{u=1}$ and $\tilde V_{u=0}$) with a continuous function of $U$ that outputs
an ordering $\tilde V_{u}$. If $T$ is sampled conditional on multiple
structured variables, then we need a function that maps from those variables
to an ordering over the vocabulary. In such a setting, we need one or more
$\tau$ hyperparameters that control how sensitive this function is to changes
in one or more structured variables.

The DAG in Figure~\ref{fig:dag_differential} also changes the assumptions
for the causal methods we consider. The Propensity and Representation methods,
like any matching estimator, requires matching only on \emph{pre-treatment}
covariates; variables that are non-descendants of the treatment $A$.
Matching on post-treatment variables can introduce significant
bias~\citep{rosenbaum1984consequences,stuart2010matching}.
If the text data is influenced by both $U$ and $A$, it cannot be easily used
for matching. Similarly, for the IPW model (or an outcome model),
if the text is a collider (descendant of both $A$ and $Y$), conditioning
on it may introduce bias~\citep{greenland2003quantifying}.

\begin{sloppy}
\binoppenalty=3000
\relpenalty=3000
Within the context of the measurement error estimator,
Figure~\ref{fig:dag_differential} violates our previous assumption of
non-differential measurement
error~\citep{carroll2006measurement,wood2018challenges}.  Thus, rather than
estimating two (assuming $U$ is binary) marginal error rates
$\prob{U^*=1}{U=0}$ and $\prob{U^*=0}{U=1}$, we must estimate several
conditional error rates of the form $\prob{U^*=u'}{U=u,A=a,C=c,Y=y}$.
Estimating such error rates requires data on the full joint $\prob{U, C, A, Y,
T}{}$ which, as discussed in \S~\ref{sec:n_labeled}, reduces the efficacy of
these methods compared to simpler approaches that ignore the text data
entirely.
\end{sloppy}

In the DAG in Figure~\ref{fig:text_treatment_outcome}, the text $T$ can be
seen as a treatment or an outcome; $p(T(a))$ is the counterfactual
distribution over $T$ if we intervene on $A$, and $p(Y(t))$
is the counterfactual distribution over $Y$ if we
intervene on $T$. Because our framework currently does not support
sampling structured variables conditional on the text, we cannot
sample from $\prob{Y}{T, U, C}{}$. The causal estimators we consider
do not make the necessary assumptions to estimate the high-dimensional effects
of $A$ on $T$ or of $T$ on $Y$~\citep{nabi2017semiparametric,egami2018make}.

\subsection{Other Language Models} \label{subsec:newer_lms}

Recent years have seen an explosion in both the frequency and size
of neural language models~\citep{bender2021dangers}.
While the only such model we have considered is a compressed version
of GPT-2~\citep{sanh2019distilbert,radford2019language}, our framework
for adding causal effects can be easily extended to new language models
such as GPT-3 or Switch-C~\citep{brown2020language,fedus2021switch}.
All our approach assumes is that the model takes as input an initial context
and then, for each word, outputs a distribution over the vocabulary. Our causal
effects simply adjust the distribution over context inputs and the distribution
over the word logits. 

Other work on language modeling has focused on \emph{controllable} text
generation which can produce sentences that follow a specified
style~\citep{xu2020controllable,keskar2019ctrl,kedzie2020controllable}.  For
example, the approach from \citet{dathathri2019plug} specifies topic (e.g.
politics) and a sentiment (e.g.  negative) which guides the text generation.
Such an approach could help generate synthetic datasets which are more
domain-specific (see \S~\ref{subsec:realistic_text}). In any future work
analyzing synthetic text generated from large-scale language models,
researchers should be careful to examine how such models learn and reproduce
societal biases encoded in the training
data~\citep{sheng2019woman,bender2021dangers}.

\subsection{Better Estimators} \label{subsec:better_estimators}

We have mentioned in \S~\ref{sec:n_labeled} that future work should consider
multiply-robust estimators with better asymptotic properties. Our evaluations
could also be extended by implementing more flexible (e.g. neural) nuisance
models that capture relationship between the structured variables and the text.
\citet{veitch2020adapting} proposed causal methods that leverage existing text
embeddings which have been widely successful in many predictive tasks. Such
neural models may require new assumptions -- such as with respect to
smoothness~\citep{farrell2021deep} -- but have demonstrated empirical
performance greatly surpassing that of the bag-of-words logistic regression
models we have considered~\citep{rajpurkar2016squad}. Such neural models often
require large datasets for training or pre-training, and in our initial
experiments, such models did not outperform logistic regression on our small
datasets. Future work could combine pre-training on large
datasets~\citep{lee2020biobert} with fine-tuning on our small
datasets~\citep{jin2019pubmedqa}.  We could also compare against stronger
baselines that ignore the text but leverage all available data, such as the
estimator of \citet{yang2020combining} which combines both a small dataset that
includes the unobserved confounder and a large dataset that does not. Such an
estimator should outperform the $\prob{U, C, A, Y}{}$ baseline we considered in
\S~\ref{sec:n_labeled} by leveraging the additional data that does not contain
$U$.

\subsection{More realistic DGPs} \label{subsec:realistic_text}

Our synthetic DGPs enable new evaluations for causal methods for text, but
synthetic data in general is not without its inherent limitations. One barrier
that prevents generalizability of results on synthetic data to real-world data
is that often synthetic DGPs are explicitly designed to demonstrate the utility
of a proposed method, and thus other assumptions that could expose the method's
flaws may be ignored by the creator~\citep{gentzel2019case}.  While our
framework addresses some of these concerns by making it easy to randomize the
DGP parameterization and enabling extensions to new language models, there is
more that can be done.  \citet{gentzel2019case} suggests semi-synthetic
datasets that, for example, use $\prob{U, C}{}$ data from a real-world study
and then sample $\prob{A, Y}{U, C}$ synthetically so the causal effects are
known~\citep{dorie2019automated,shimoni2018benchmarking}. While our framework
could adopt this approach and use empirical $\prob{U, C}{}$ data, if we use
empirical text data we lose any knowledge of the causal relationships between
text and structured variables.

Within the synthetic framework we have proposed, there are many ways to make
our synthetic DGPs more realistic for applications to specific domain areas.
We have used EHR data and clinical notes as a motivating example throughout,
but our DGPs are unrelated to such applications. Suppose we have an EHR dataset
with physiological measurements and clinical notes. If we want to conduct a
retrospective causal analysis using text, we might first develop a synthetic
DGP that tries to approximate the empirical dataset~\citep{neal2020realcause}.
To adapt the synthetic DGPs from this work to this application, we might
consider using a language model fine-tuned on clinical
notes~\citep{lee2020biobert} or adapted to the complex vocabulary and style of
the domain~\citep{ruch2003using,melamud2019towards,boag2016towards,
choi2017generating}.  If our clinical data has a structured variable $U$ that
we believe influences the text $T$, we might incorporate controllable
generation techniques to parameterize $\prob{T}{U}$ more realistically, for
example by choosing a vocabulary preference $\tilde V_{u}$ that reflect which
words are more commonly used when describing patients with different values of
$U$. Such adaptations could make inferences drawn from synthetic data more
robust or make evaluations more interpretable to domain experts.

\section{Conclusions}

Our experiments demonstrate the importance of accurate assumptions in a causal
analysis. All four causal methods can control for unobserved confounding in a
trivial text generation setting, but as our generative $\prob{T}{U}$ increases in
complexity, the implicit assumptions of the matching and IPW methods render
them biased. Although the matching and IPW methods use the same
$\prob{A}{C, T}$ propensity score model, the matching approaches work are
superior in the trivial and LDA settings. Even though the trained models are
identical, the underlying assumptions are different.
Because it requires additional data, the measurement error approach is able to
make fewer assumptions, remaining effective as long as its
$\prob{U}{T}$ classifier is accurate. These results do not imply that text
matching and IPW methods {\it cannot} control for unobserved confounding, but
rather that we should be cautious and clear about what assumptions we make
about our models and the underlying DGP. Evaluating on synthetic data can help
clarify these assumptions.

As NLP research furthers the state-of-the-art in predictive modeling, such
tools offer the potential to influence human decision-making and guide our
understanding of the world. Such models rely on assumptions that may be
irrelevant for a supervised learning benchmark and yet essential to any
real-world application. Explicitly adopting a causal inference perspective on
natural language datasets can help enable inferences that are robust to
confounding or other biases. We hope our evaluation framework and released
code will support further research in these directions.

\clearpage

\bibliography{paper,references}

\begin{thebibliography}{81}
\providecommand{\natexlab}[1]{#1}
\providecommand{\url}[1]{\texttt{#1}}
\expandafter\ifx\csname urlstyle\endcsname\relax
  \providecommand{\doi}[1]{doi: #1}\else
  \providecommand{\doi}{doi: \begingroup \urlstyle{rm}\Url}\fi

\bibitem[Afzal et~al.(2018)Afzal, Mallipeddi, Sohn, Liu, Chaudhry, Scott,
  Kullo, and Arruda-Olson]{afzal2018natural}
Naveed Afzal, Vishnu~Priya Mallipeddi, Sunghwan Sohn, Hongfang Liu, Rajeev
  Chaudhry, Christopher~G Scott, Iftikhar~J Kullo, and Adelaide~M Arruda-Olson.
\newblock Natural language processing of clinical notes for identification of
  critical limb ischemia.
\newblock \emph{International journal of medical informatics}, 111:\penalty0
  83--89, 2018.

\bibitem[Anderson and Magruder(2017)]{anderson2017split}
Michael~L Anderson and Jeremy Magruder.
\newblock Split-sample strategies for avoiding false discoveries.
\newblock Technical report, National Bureau of Economic Research, 2017.

\bibitem[Bang and Robins(2005)]{bang2005doubly}
Heejung Bang and James~M Robins.
\newblock Doubly robust estimation in missing data and causal inference models.
\newblock \emph{Biometrics}, 61\penalty0 (4):\penalty0 962--973, 2005.

\bibitem[Belinkov and Bisk(2018)]{belinkov2018synthetic}
Yonatan Belinkov and Yonatan Bisk.
\newblock Synthetic and natural noise both break neural machine translation.
\newblock In \emph{International Conference on Learning Representations}, 2018.

\bibitem[Bender et~al.(2021)Bender, Gebru, McMillan-Major, and
  Shmitchell]{bender2021dangers}
Emily~M. Bender, Timnit Gebru, Angelina McMillan-Major, and Shmargaret
  Shmitchell.
\newblock On the dangers of stochastic parrots: Can language models be too big?
\newblock In \emph{Proceedings of FAccT}, 2021.

\bibitem[Blei and McAuliffe(2007)]{blei2007supervised}
David~M Blei and Jon~D McAuliffe.
\newblock Supervised topic models.
\newblock In \emph{Proceedings of the 20th International Conference on Neural
  Information Processing Systems}, pages 121--128, 2007.

\bibitem[Blei et~al.(2003)Blei, Ng, and Jordan]{blei2003latent}
David~M Blei, Andrew~Y Ng, and Michael~I Jordan.
\newblock Latent dirichlet allocation.
\newblock \emph{Journal of machine Learning research}, 3\penalty0
  (Jan):\penalty0 993--1022, 2003.

\bibitem[Boag et~al.(2016)Boag, Naumann, and Szolovits]{boag2016towards}
Willie Boag, Tristan Naumann, and Peter Szolovits.
\newblock Towards the creation of a large corpus of synthetically-identified
  clinical notes.
\newblock In \emph{Machine Learning for Health Workshop at NeurIPS}, 2016.

\bibitem[Bodnar et~al.(2014)Bodnar, Simhan, Catov, Roberts, Platt, Diesel, and
  Klebanoff]{bodnar2014maternal}
Lisa~M Bodnar, Hyagriv~N Simhan, Janet~M Catov, James~M Roberts, Robert~W
  Platt, Jill~C Diesel, and Mark~A Klebanoff.
\newblock Maternal vitamin d status and the risk of mild and severe
  preeclampsia.
\newblock \emph{Epidemiology (Cambridge, Mass.)}, 25\penalty0 (2):\penalty0
  207, 2014.

\bibitem[Brown et~al.(2020)Brown, Mann, Ryder, Subbiah, Kaplan, Dhariwal,
  Neelakantan, Shyam, Sastry, Askell, Agarwal, Herbert, Henighan, Child,
  Ramesh, Ziegler, Wu, Winter, Hesse, Chen, Sigler, Litwin, Gray, Chess, Clark,
  Berner, McCandlish, Radford, Sutskever, and Amodei]{brown2020language}
Tom Brown, Benjamin Mann, Nick Ryder, Melanie Subbiah, Jared Kaplan, Prafulla
  Dhariwal, Arvind Neelakantan, Pranav Shyam, Girish Sastry, Amanda Askell,
  Sandhini Agarwal, Gretchen Herbert, Ariel and-Voss~Krueger, Tom Henighan,
  Rewon Child, Aditya Ramesh, Daniel Ziegler, Jeffrey Wu, Clemens Winter, Chris
  Hesse, Mark Chen, Eric Sigler, Mateusz Litwin, Scott Gray, Benjamin Chess,
  Jack Clark, Christopher Berner, Sam McCandlish, Alec Radford, Ilya Sutskever,
  and Dario Amodei.
\newblock Language models are few-shot learners.
\newblock In \emph{Advances in Neural Information Processing Systems}, 2020.

\bibitem[Carroll et~al.(2006)Carroll, Ruppert, Stefanski, and
  Crainiceanu]{carroll2006measurement}
Raymond~J Carroll, David Ruppert, Leonard~A Stefanski, and Ciprian~M
  Crainiceanu.
\newblock \emph{Measurement error in nonlinear models: a modern perspective}.
\newblock CRC press, 2006.

\bibitem[Char et~al.(2018)Char, Shah, and Magnus]{char2018implementing}
Danton~S Char, Nigam~H Shah, and David Magnus.
\newblock Implementing machine learning in health care -- addressing ethical
  challenges.
\newblock \emph{The New England journal of medicine}, 378\penalty0
  (11):\penalty0 981, 2018.

\bibitem[Chen and Asch(2017)]{chen2017machine}
Jonathan~H Chen and Steven~M Asch.
\newblock Machine learning and prediction in medicine -- beyond the peak of
  inflated expectations.
\newblock \emph{The New England journal of medicine}, 376\penalty0
  (26):\penalty0 2507, 2017.

\bibitem[Chernozhukov et~al.(2016)Chernozhukov, Chetverikov, Demirer, Duflo,
  Hansen, and Newey]{chernozhukov2016double}
Victor Chernozhukov, Denis Chetverikov, Mert Demirer, Esther Duflo, Christian
  Hansen, and Whitney~K Newey.
\newblock Double machine learning for treatment and causal parameters.
\newblock Technical report, cemmap working paper, 2016.

\bibitem[Choi et~al.(2017)Choi, Biswal, Malin, Duke, Stewart, and
  Sun]{choi2017generating}
Edward Choi, Siddharth Biswal, Bradley Malin, Jon Duke, Walter~F Stewart, and
  Jimeng Sun.
\newblock Generating multi-label discrete patient records using generative
  adversarial networks.
\newblock In \emph{Machine learning for healthcare conference}, pages 286--305.
  PMLR, 2017.

\bibitem[Dathathri et~al.(2019)Dathathri, Madotto, Lan, Hung, Frank, Molino,
  Yosinski, and Liu]{dathathri2019plug}
Sumanth Dathathri, Andrea Madotto, Janice Lan, Jane Hung, Eric Frank, Piero
  Molino, Jason Yosinski, and Rosanne Liu.
\newblock Plug and play language models: A simple approach to controlled text
  generation.
\newblock In \emph{International Conference on Learning Representations}, 2019.

\bibitem[Deng et~al.(2009)Deng, Dong, Socher, Li, Li, and
  Fei-Fei]{deng2009imagenet}
Jia Deng, Wei Dong, Richard Socher, Li-Jia Li, Kai Li, and Li~Fei-Fei.
\newblock Imagenet: A large-scale hierarchical image database.
\newblock In \emph{2009 IEEE conference on computer vision and pattern
  recognition}, pages 248--255. Ieee, 2009.

\bibitem[Dorie et~al.(2019)Dorie, Hill, Shalit, Scott, Cervone,
  et~al.]{dorie2019automated}
Vincent Dorie, Jennifer Hill, Uri Shalit, Marc Scott, Dan Cervone, et~al.
\newblock Automated versus do-it-yourself methods for causal inference: Lessons
  learned from a data analysis competition.
\newblock \emph{Statistical Science}, 34\penalty0 (1):\penalty0 43--68, 2019.

\bibitem[Egami et~al.(2018)Egami, Fong, Grimmer, Roberts, and
  Stewart]{egami2018make}
Naoki Egami, Christian~J Fong, Justin Grimmer, Margaret~E Roberts, and
  Brandon~M Stewart.
\newblock How to make causal inferences using texts.
\newblock \emph{arXiv preprint arXiv:1802.02163}, 2018.

\bibitem[Elman(1990)]{elman1990finding}
Jeffrey~L Elman.
\newblock Finding structure in time.
\newblock \emph{Cognitive science}, 14\penalty0 (2):\penalty0 179--211, 1990.

\bibitem[Farrell et~al.(2021)Farrell, Liang, and Misra]{farrell2021deep}
Max~H Farrell, Tengyuan Liang, and Sanjog Misra.
\newblock Deep neural networks for estimation and inference.
\newblock \emph{Econometrica}, 89\penalty0 (1):\penalty0 181--213, 2021.

\bibitem[Fedus et~al.(2021)Fedus, Zoph, and Shazeer]{fedus2021switch}
William Fedus, Barret Zoph, and Noam Shazeer.
\newblock Switch transformers: Scaling to trillion parameter models with simple
  and efficient sparsity.
\newblock \emph{arXiv preprint arXiv:2101.03961}, 2021.

\bibitem[Gentzel et~al.(2019)Gentzel, Garant, and Jensen]{gentzel2019case}
Amanda Gentzel, Dan Garant, and David Jensen.
\newblock The case for evaluating causal models using interventional measures
  and empirical data.
\newblock In \emph{Advances in Neural Information Processing Systems}, pages
  11722--11732, 2019.

\bibitem[Greenland(2003)]{greenland2003quantifying}
Sander Greenland.
\newblock Quantifying biases in causal models: classical confounding vs
  collider-stratification bias.
\newblock \emph{Epidemiology}, pages 300--306, 2003.

\bibitem[Hahn et~al.(2019)Hahn, Dorie, and Murray]{hahn2019atlantic}
P~Richard Hahn, Vincent Dorie, and Jared~S Murray.
\newblock Atlantic causal inference conference (acic) data analysis challenge
  2017.
\newblock \emph{arXiv preprint arXiv:1905.09515}, 2019.

\bibitem[Hashimoto et~al.(2019)Hashimoto, Zhang, and
  Liang]{hashimoto2019unifying}
Tatsunori Hashimoto, Hugh Zhang, and Percy Liang.
\newblock Unifying human and statistical evaluation for natural language
  generation.
\newblock In \emph{Proceedings of the 2019 Conference of the North American
  Chapter of the Association for Computational Linguistics: Human Language
  Technologies, Volume 1 (Long and Short Papers)}, pages 1689--1701, 2019.

\bibitem[Hill et~al.(2011)Hill, Weiss, and Zhai]{hill2011challenges}
Jennifer Hill, Christopher Weiss, and Fuhua Zhai.
\newblock Challenges with propensity score strategies in a high-dimensional
  setting and a potential alternative.
\newblock \emph{Multivariate Behavioral Research}, 46\penalty0 (3):\penalty0
  477--513, 2011.

\bibitem[Jensen et~al.(2019)]{jensen2019comment}
David Jensen et~al.
\newblock Comment: Strengthening empirical evaluation of causal inference
  methods.
\newblock \emph{Statistical Science}, 34\penalty0 (1):\penalty0 77--81, 2019.

\bibitem[Jin et~al.(2019)Jin, Dhingra, Liu, Cohen, and Lu]{jin2019pubmedqa}
Qiao Jin, Bhuwan Dhingra, Zhengping Liu, William Cohen, and Xinghua Lu.
\newblock Pubmedqa: A dataset for biomedical research question answering.
\newblock In \emph{Proceedings of the 2019 Conference on Empirical Methods in
  Natural Language Processing and the 9th International Joint Conference on
  Natural Language Processing (EMNLP-IJCNLP)}, pages 2567--2577, 2019.

\bibitem[Johansson et~al.(2016)Johansson, Shalit, and
  Sontag]{johansson2016learning}
Fredrik Johansson, Uri Shalit, and David Sontag.
\newblock Learning representations for counterfactual inference.
\newblock In \emph{International conference on machine learning}, pages
  3020--3029, 2016.

\bibitem[Kedzie and McKeown(2020)]{kedzie2020controllable}
Chris Kedzie and Kathleen McKeown.
\newblock Controllable meaning representation to text generation: Linearization
  and data augmentation strategies.
\newblock In \emph{Proceedings of the 2020 Conference on Empirical Methods in
  Natural Language Processing (EMNLP)}, pages 5160--5185, Online, November
  2020. Association for Computational Linguistics.
\newblock \doi{10.18653/v1/2020.emnlp-main.419}.

\bibitem[Keith et~al.(2020)Keith, Jensen, and O{'}Connor]{keith2020text}
Katherine Keith, David Jensen, and Brendan O{'}Connor.
\newblock Text and causal inference: A review of using text to remove
  confounding from causal estimates.
\newblock In \emph{Proceedings of the 58th Annual Meeting of the Association
  for Computational Linguistics}, pages 5332--5344, Online, July 2020.
  Association for Computational Linguistics.
\newblock \doi{10.18653/v1/2020.acl-main.474}.
\newblock URL \url{https://www.aclweb.org/anthology/2020.acl-main.474}.

\bibitem[Keskar et~al.(2019)Keskar, McCann, Varshney, Xiong, and
  Socher]{keskar2019ctrl}
Nitish~Shirish Keskar, Bryan McCann, Lav~R Varshney, Caiming Xiong, and Richard
  Socher.
\newblock Ctrl: A conditional transformer language model for controllable
  generation.
\newblock \emph{arXiv preprint arXiv:1909.05858}, 2019.

\bibitem[Khayrallah and Koehn(2018)]{khayrallah2018impact}
Huda Khayrallah and Philipp Koehn.
\newblock On the impact of various types of noise on neural machine
  translation.
\newblock In \emph{Proceedings of the 2nd Workshop on Neural Machine
  Translation and Generation}, pages 74--83, 2018.

\bibitem[Kim and O{'}Neill-Brown(2019)]{kim-oneill-brown-2019-improving}
Jungi Kim and Patricia O{'}Neill-Brown.
\newblock Improving {A}merican {S}ign {L}anguage recognition with synthetic
  data.
\newblock In \emph{Proceedings of Machine Translation Summit XVII Volume 1:
  Research Track}, pages 151--161, Dublin, Ireland, August 2019. European
  Association for Machine Translation.
\newblock URL \url{https://www.aclweb.org/anthology/W19-6615}.

\bibitem[Lee et~al.(2011)Lee, Lessler, and Stuart]{lee2011weight}
Brian~K Lee, Justin Lessler, and Elizabeth~A Stuart.
\newblock Weight trimming and propensity score weighting.
\newblock \emph{PloS one}, 6\penalty0 (3), 2011.

\bibitem[Lee et~al.(2020)Lee, Yoon, Kim, Kim, Kim, So, and
  Kang]{lee2020biobert}
Jinhyuk Lee, Wonjin Yoon, Sungdong Kim, Donghyeon Kim, Sunkyu Kim, Chan~Ho So,
  and Jaewoo Kang.
\newblock Biobert: a pre-trained biomedical language representation model for
  biomedical text mining.
\newblock \emph{Bioinformatics}, 36\penalty0 (4):\penalty0 1234--1240, 2020.

\bibitem[Liu et~al.(2018)Liu, Zhang, and Razavian]{liu2018deep}
Jingshu Liu, Zachariah Zhang, and Narges Razavian.
\newblock Deep ehr: Chronic disease prediction using medical notes.
\newblock In \emph{Machine Learning for Healthcare Conference}, pages 440--464,
  2018.

\bibitem[Liu et~al.(2019)Liu, Bates, Wiens, and Shah]{liu2019number}
Vincent~X Liu, David~W Bates, Jenna Wiens, and Nigam~H Shah.
\newblock The number needed to benefit: estimating the value of predictive
  analytics in healthcare.
\newblock \emph{Journal of the American Medical Informatics Association},
  26\penalty0 (12):\penalty0 1655--1659, 2019.

\bibitem[Low et~al.(2016)Low, Gallego, and Shah]{low2016comparing}
Yen~Sia Low, Blanca Gallego, and Nigam~Haresh Shah.
\newblock Comparing high-dimensional confounder control methods for rapid
  cohort studies from electronic health records.
\newblock \emph{Journal of comparative effectiveness research}, 5\penalty0
  (2):\penalty0 179--192, 2016.

\bibitem[McCaffrey et~al.(2004)McCaffrey, Ridgeway, and
  Morral]{mccaffrey2004propensity}
Daniel~F McCaffrey, Greg Ridgeway, and Andrew~R Morral.
\newblock Propensity score estimation with boosted regression for evaluating
  causal effects in observational studies.
\newblock \emph{Psychological methods}, 9\penalty0 (4):\penalty0 403, 2004.

\bibitem[McVeigh et~al.(2016)McVeigh, Newton-Dame, Chan, Thorpe, Schreibstein,
  Tatem, Chernov, Lurie-Moroni, and Perlman]{mcveigh2016can}
Katharine~H McVeigh, Remle Newton-Dame, Pui~Ying Chan, Lorna~E Thorpe, Lauren
  Schreibstein, Kathleen~S Tatem, Claudia Chernov, Elizabeth Lurie-Moroni, and
  Sharon~E Perlman.
\newblock Can electronic health records be used for population health
  surveillance? validating population health metrics against established survey
  data.
\newblock \emph{eGEMs}, 4\penalty0 (1), 2016.

\bibitem[Melamud and Shivade(2019)]{melamud2019towards}
Oren Melamud and Chaitanya Shivade.
\newblock Towards automatic generation of shareable synthetic clinical notes
  using neural language models.
\newblock In \emph{Proceedings of the 2nd Clinical Natural Language Processing
  Workshop}, pages 35--45, 2019.

\bibitem[Meystre et~al.(2008)Meystre, Savova, Kipper-Schuler, and
  Hurdle]{meystre2008extracting}
St{\'e}phane~M Meystre, Guergana~K Savova, Karin~C Kipper-Schuler, and John~F
  Hurdle.
\newblock Extracting information from textual documents in the electronic
  health record: a review of recent research.
\newblock \emph{Yearbook of medical informatics}, 17\penalty0 (01):\penalty0
  128--144, 2008.

\bibitem[Mozer et~al.(2018)Mozer, Miratrix, Kaufman, and
  Anastasopoulos]{mozer2018matching}
Reagan Mozer, Luke Miratrix, Aaron~Russell Kaufman, and L~Jason Anastasopoulos.
\newblock Matching with text data: An experimental evaluation of methods for
  matching documents and of measuring match quality.
\newblock \emph{arXiv preprint arXiv:1801.00644}, 2018.

\bibitem[Nabi et~al.(2017)Nabi, McNutt, and Shpitser]{nabi2017semiparametric}
Razieh Nabi, Todd McNutt, and Ilya Shpitser.
\newblock Semiparametric causal sufficient dimension reduction of high
  dimensional treatments.
\newblock \emph{arXiv preprint arXiv:1710.06727}, 2017.

\bibitem[Neal et~al.(2020)Neal, Huang, and Raghupathi]{neal2020realcause}
Brady Neal, Chin-Wei Huang, and Sunand Raghupathi.
\newblock Realcause: Realistic causal inference benchmarking.
\newblock \emph{arXiv preprint arXiv:2011.15007}, 2020.

\bibitem[Patki et~al.(2016)Patki, Wedge, and
  Veeramachaneni]{patki2016synthetic}
Neha Patki, Roy Wedge, and Kalyan Veeramachaneni.
\newblock The synthetic data vault.
\newblock In \emph{2016 IEEE International Conference on Data Science and
  Advanced Analytics (DSAA)}, pages 399--410. IEEE, 2016.

\bibitem[Pearl(2009)]{pearl2009causality}
Judea Pearl.
\newblock \emph{Causality}.
\newblock Cambridge University Press, 2009.

\bibitem[Pearl(2010)]{pearl2010measurement}
Judea Pearl.
\newblock On measurement bias in causal inference.
\newblock In \emph{Proceedings of the Twenty-Sixth Conference on Uncertainty in
  Artificial Intelligence}, pages 425--432, 2010.

\bibitem[Pearl and Mackenzie(2018)]{pearl2018book}
Judea Pearl and Dana Mackenzie.
\newblock \emph{The book of why: the new science of cause and effect}.
\newblock Basic Books, 2018.

\bibitem[Radford et~al.(2019)Radford, Wu, Child, Luan, Amodei, and
  Sutskever]{radford2019language}
Alec Radford, Jeffrey Wu, Rewon Child, David Luan, Dario Amodei, and Ilya
  Sutskever.
\newblock Language models are unsupervised multitask learners.
\newblock \emph{OpenAI Blog}, 1\penalty0 (8):\penalty0 9, 2019.

\bibitem[Rajkomar et~al.(2018)Rajkomar, Oren, Chen, Dai, Hajaj, Hardt, Liu,
  Liu, Marcus, Sun, et~al.]{rajkomar2018scalable}
Alvin Rajkomar, Eyal Oren, Kai Chen, Andrew~M Dai, Nissan Hajaj, Michaela
  Hardt, Peter~J Liu, Xiaobing Liu, Jake Marcus, Mimi Sun, et~al.
\newblock Scalable and accurate deep learning with electronic health records.
\newblock \emph{NPJ Digital Medicine}, 1\penalty0 (1):\penalty0 18, 2018.

\bibitem[Rajpurkar et~al.(2016)Rajpurkar, Zhang, Lopyrev, and
  Liang]{rajpurkar2016squad}
Pranav Rajpurkar, Jian Zhang, Konstantin Lopyrev, and Percy Liang.
\newblock Squad: 100,000+ questions for machine comprehension of text.
\newblock In \emph{Proceedings of the 2016 Conference on Empirical Methods in
  Natural Language Processing}, pages 2383--2392, 2016.

\bibitem[Roberts et~al.(2018)Roberts, Stewart, and
  Nielsen]{roberts2018adjusting}
Margaret~E Roberts, Brandon~M Stewart, and Richard~A Nielsen.
\newblock Adjusting for confounding with text matching.
\newblock \emph{American Journal of Political Science}, 2018.

\bibitem[Rosenbaum(1984)]{rosenbaum1984consequences}
Paul~R Rosenbaum.
\newblock The consequences of adjustment for a concomitant variable that has
  been affected by the treatment.
\newblock \emph{Journal of the Royal Statistical Society: Series A (General)},
  147\penalty0 (5):\penalty0 656--666, 1984.

\bibitem[Rosenbaum and Rubin(1983)]{rosenbaum1983central}
Paul~R Rosenbaum and Donald~B Rubin.
\newblock The central role of the propensity score in observational studies for
  causal effects.
\newblock \emph{Biometrika}, 70\penalty0 (1):\penalty0 41--55, 1983.

\bibitem[Rosenbloom et~al.(2011)Rosenbloom, Denny, Xu, Lorenzi, Stead, and
  Johnson]{rosenbloom2011data}
S~Trent Rosenbloom, Joshua~C Denny, Hua Xu, Nancy Lorenzi, William~W Stead, and
  Kevin~B Johnson.
\newblock Data from clinical notes: a perspective on the tension between
  structure and flexible documentation.
\newblock \emph{Journal of the American Medical Informatics Association},
  18\penalty0 (2):\penalty0 181--186, 2011.

\bibitem[Ruch et~al.(2003)Ruch, Baud, and Geissb{\"u}hler]{ruch2003using}
Patrick Ruch, Robert Baud, and Antoine Geissb{\"u}hler.
\newblock Using lexical disambiguation and named-entity recognition to improve
  spelling correction in the electronic patient record.
\newblock \emph{Artificial intelligence in medicine}, 29\penalty0
  (1-2):\penalty0 169--184, 2003.

\bibitem[Sanh et~al.(2019)Sanh, Debut, Chaumond, and Wolf]{sanh2019distilbert}
Victor Sanh, Lysandre Debut, Julien Chaumond, and Thomas Wolf.
\newblock Distilbert, a distilled version of bert: smaller, faster, cheaper and
  lighter.
\newblock \emph{arXiv preprint arXiv:1910.01108v4}, 2019.

\bibitem[Savova et~al.(2010)Savova, Masanz, Ogren, Zheng, Sohn, Kipper-Schuler,
  and Chute]{savova2010mayo}
Guergana~K Savova, James~J Masanz, Philip~V Ogren, Jiaping Zheng, Sunghwan
  Sohn, Karin~C Kipper-Schuler, and Christopher~G Chute.
\newblock {Mayo clinical Text Analysis and Knowledge Extraction System
  (cTAKES): architecture, component evaluation and applications}.
\newblock \emph{JAMIA}, 17\penalty0 (5):\penalty0 507--513, 09 2010.
\newblock ISSN 1067-5027.
\newblock \doi{10.1136/jamia.2009.001560}.
\newblock URL \url{https://doi.org/10.1136/jamia.2009.001560}.

\bibitem[Sheng et~al.(2019)Sheng, Chang, Natarajan, and Peng]{sheng2019woman}
Emily Sheng, Kai-Wei Chang, Prem Natarajan, and Nanyun Peng.
\newblock The woman worked as a babysitter: On biases in language generation.
\newblock In \emph{Proceedings of the 2019 Conference on Empirical Methods in
  Natural Language Processing and the 9th International Joint Conference on
  Natural Language Processing (EMNLP-IJCNLP)}, pages 3398--3403, 2019.

\bibitem[Shimoni et~al.(2018)Shimoni, Yanover, Karavani, and
  Goldschmnidt]{shimoni2018benchmarking}
Yishai Shimoni, Chen Yanover, Ehud Karavani, and Yaara Goldschmnidt.
\newblock Benchmarking framework for performance-evaluation of causal inference
  analysis.
\newblock \emph{arXiv preprint arXiv:1802.05046}, 2018.

\bibitem[Shpitser and Pearl(2006)]{shpitser2006identification}
Ilya Shpitser and Judea Pearl.
\newblock Identification of joint interventional distributions in recursive
  semi-markovian causal models.
\newblock In \emph{21st National Conference on Artificial Intelligence and the
  18th Innovative Applications of Artificial Intelligence Conference,
  AAAI-06/IAAI-06}, pages 1219--1226, 2006.

\bibitem[Silva et~al.(2008)Silva, Coolman, Steegers, Jaddoe, Moll, Hofman,
  Mackenbach, and Raat]{silva2008low}
Lindsay~M Silva, Marianne Coolman, Eric~AP Steegers, Vincent~WV Jaddoe,
  Henriette~A Moll, Albert Hofman, Johan~P Mackenbach, and Hein Raat.
\newblock Low socioeconomic status is a risk factor for preeclampsia: the
  generation r study.
\newblock \emph{Journal of hypertension}, 26\penalty0 (6):\penalty0 1200--1208,
  2008.

\bibitem[Stuart(2010)]{stuart2010matching}
Elizabeth~A Stuart.
\newblock Matching methods for causal inference: A review and a look forward.
\newblock \emph{Statistical science: a review journal of the Institute of
  Mathematical Statistics}, 25\penalty0 (1):\penalty0 1, 2010.

\bibitem[Subbaswamy et~al.(2019)Subbaswamy, Schulam, and
  Saria]{subbaswamy2019preventing}
Adarsh Subbaswamy, Peter Schulam, and Suchi Saria.
\newblock Preventing failures due to dataset shift: Learning predictive models
  that transport.
\newblock In \emph{The 22nd International Conference on Artificial Intelligence
  and Statistics}, pages 3118--3127, 2019.

\bibitem[Vansteelandt et~al.(2008)Vansteelandt, VanderWeele, Tchetgen, and
  Robins]{vansteelandt2008multiply}
Stijn Vansteelandt, Tyler~J VanderWeele, Eric~J Tchetgen, and James~M Robins.
\newblock Multiply robust inference for statistical interactions.
\newblock \emph{Journal of the American Statistical Association}, 103\penalty0
  (484):\penalty0 1693--1704, 2008.

\bibitem[Veitch et~al.(2020)Veitch, Sridhar, and Blei]{veitch2020adapting}
Victor Veitch, Dhanya Sridhar, and David Blei.
\newblock Adapting text embeddings for causal inference.
\newblock In \emph{Conference on Uncertainty in Artificial Intelligence}, pages
  919--928. PMLR, 2020.

\bibitem[Wallach(2006)]{wallach2006topic}
Hanna~M Wallach.
\newblock Topic modeling: beyond bag-of-words.
\newblock In \emph{ICML}, pages 977--984, 2006.

\bibitem[Wang and Eisner(2018)]{wang-eisner-2018-synthetic}
Dingquan Wang and Jason Eisner.
\newblock Synthetic data made to order: The case of parsing.
\newblock In \emph{Proceedings of the 2018 Conference on Empirical Methods in
  Natural Language Processing}, pages 1325--1337, Brussels, Belgium,
  October-November 2018. Association for Computational Linguistics.
\newblock \doi{10.18653/v1/D18-1163}.
\newblock URL \url{https://www.aclweb.org/anthology/D18-1163}.

\bibitem[Wang and Culotta(2019)]{wang2019words}
Zhao Wang and Aron Culotta.
\newblock When do words matter? understanding the impact of lexical choice on
  audience perception using individual treatment effect estimation.
\newblock In \emph{Proceedings of the AAAI Conference on Artificial
  Intelligence}, volume~33, pages 7233--7240, 2019.

\bibitem[Weld et~al.(2020)Weld, West, Glenski, Arbour, Rossi, and
  Althoff]{weld2020adjusting}
Galen Weld, Peter West, Maria Glenski, David Arbour, Ryan Rossi, and Tim
  Althoff.
\newblock Adjusting for confounders with text: Challenges and an empirical
  evaluation framework for causal inference.
\newblock \emph{arXiv preprint arXiv:2009.09961}, 2020.

\bibitem[Wendling et~al.(2018)Wendling, Jung, Callahan, Schuler, Shah, and
  Gallego]{wendling2018comparing}
T~Wendling, K~Jung, A~Callahan, A~Schuler, NH~Shah, and B~Gallego.
\newblock Comparing methods for estimation of heterogeneous treatment effects
  using observational data from health care databases.
\newblock \emph{Statistics in medicine}, 37\penalty0 (23):\penalty0 3309--3324,
  2018.

\bibitem[Winata et~al.(2019)Winata, Madotto, Wu, and
  Fung]{winata-etal-2019-code}
Genta~Indra Winata, Andrea Madotto, Chien-Sheng Wu, and Pascale Fung.
\newblock Code-switched language models using neural based synthetic data from
  parallel sentences.
\newblock In \emph{Proceedings of the 23rd Conference on Computational Natural
  Language Learning (CoNLL)}, pages 271--280, Hong Kong, China, November 2019.
  Association for Computational Linguistics.
\newblock \doi{10.18653/v1/K19-1026}.
\newblock URL \url{https://www.aclweb.org/anthology/K19-1026}.

\bibitem[Wood-Doughty et~al.(2018)Wood-Doughty, Shpitser, and
  Dredze]{wood2018challenges}
Zach Wood-Doughty, Ilya Shpitser, and Mark Dredze.
\newblock Challenges of using text classifiers for causal inference.
\newblock In \emph{Proceedings of the 2018 Conference on Empirical Methods in
  Natural Language Processing}, pages 4586--4598, Brussels, Belgium, 2018.
  Association for Computational Linguistics.
\newblock \doi{10.18653/v1/D18-1488}.

\bibitem[Wu et~al.(2013)Wu, Chang, Robson, Jackson, Chen, Hayes, and
  Stewart]{wu2013evaluation}
Chia-Yi Wu, Chin-Kuo Chang, Debbie Robson, Richard Jackson, Shaw-Ji Chen,
  Richard~D Hayes, and Robert Stewart.
\newblock Evaluation of smoking status identification using electronic health
  records and open-text information in a large mental health case register.
\newblock \emph{PloS one}, 8\penalty0 (9), 2013.

\bibitem[Xu et~al.(2020)Xu, Patwary, Shoeybi, Puri, Fung, Anandkumar, and
  Catanzaro]{xu2020controllable}
Peng Xu, Mostofa Patwary, Mohammad Shoeybi, Raul Puri, Pascale Fung, Animashree
  Anandkumar, and Bryan Catanzaro.
\newblock Controllable story generation with external knowledge using
  large-scale language models.
\newblock In \emph{Proceedings of the 2020 Conference on Empirical Methods in
  Natural Language Processing (EMNLP)}, pages 2831--2845, 2020.

\bibitem[Yang and Ding(2020)]{yang2020combining}
Shu Yang and Peng Ding.
\newblock Combining multiple observational data sources to estimate causal
  effects.
\newblock \emph{Journal of the American Statistical Association}, 115\penalty0
  (531):\penalty0 1540--1554, 2020.

\bibitem[Yao et~al.(2019)Yao, Li, Li, Xue, Gao, and Zhang]{yao2019estimation}
Liuyi Yao, Sheng Li, Yaliang Li, Hongfei Xue, Jing Gao, and Aidong Zhang.
\newblock On the estimation of treatment effect with text covariates.
\newblock In \emph{Proceedings of the 28th International Joint Conference on
  Artificial Intelligence}, pages 4106--4113. AAAI Press, 2019.

\bibitem[Zheng et~al.(2011)Zheng, Hanauer, Padman, Johnson, Hussain, Ye, Zhou,
  and Diamond]{zheng2011handling}
Kai Zheng, David~A Hanauer, Rema Padman, Michael~P Johnson, Anwar~A Hussain,
  Wen Ye, Xiaomu Zhou, and Herbert~S Diamond.
\newblock Handling anticipated exceptions in clinical care: investigating
  clinician use of `exit strategies' in an electronic health records system.
\newblock \emph{Journal of the American Medical Informatics Association},
  18\penalty0 (6):\penalty0 883--889, 2011.

\end{thebibliography}

\end{document}